%% file: 0-main.tex
\documentclass[10pt,twocolumn,letterpaper]{article}

\usepackage{3dv}
\usepackage{times}
\usepackage{epsfig}
\usepackage{graphicx}
\usepackage{amsmath}
\usepackage{amssymb}
\usepackage{amsthm}
\usepackage{microtype}
\usepackage{mathtools}
\usepackage{multirow}
\usepackage{array}
\usepackage{booktabs}
\usepackage{algorithm}
\usepackage{algpseudocode}
\usepackage{varwidth} 

\usepackage[pagebackref=true,breaklinks=true,colorlinks,bookmarks=false]{hyperref}
\graphicspath{{./figures/}}

\algnewcommand\algorithmicinput{\textbf{Input:}}
\algnewcommand\INPUT{\item[\algorithmicinput]}

\threedvfinalcopy 

\def\threedvPaperID{0084} 

\usepackage{enumitem}
\setitemize[0]{leftmargin=10pt}

\newenvironment{tight_itemize}{
\begin{itemize}[leftmargin=10pt]
  \setlength{\topsep}{0pt}
  \setlength{\itemsep}{2pt}
  \setlength{\parskip}{0pt}
  \setlength{\parsep}{0pt}
}{\end{itemize}}

\DeclarePairedDelimiterX{\infdivx}[2]{}{}{%
  #1\;\delimsize\|\;#2%
}

\newcommand{\specialcell}[2][c]{%
  \begin{tabular}[#1]{@{}c@{}}#2\end{tabular}}
\newcommand{\todo}[1]{\textcolor{red}{TODO: #1}\PackageWarning{TODO:}{#1!}}

\newcommand{\ignore}[1]{}
\newcommand{\mcreconfull}[0]{Weakly supervised 3D Reconstruction with Adversarial Constraint\xspace}
\newcommand{\mcrecon}[0]{McRecon\xspace}
\newcommand{\raytracinglayer}[0]{RP-Layer\xspace}
\renewcommand{\algorithmiccomment}[1]{\bgroup\hfill//~#1\egroup}

\makeatletter
\newcounter{ALC@tempcntr}
\makeatother

\ifthreedvfinal\fi

\begin{document}

\title{\mcreconfull}

\author{JunYoung Gwak\thanks{indicates equal contributions}\\
Stanford University\\
{\tt\small jgwak@stanford.edu}
\and
Christopher B. Choy\textsuperscript{*}\\
Stanford University\\
{\tt\small chrischoy@stanford.edu}
\and
Manmohan Chandraker\\
NEC Laboratories America, Inc.\\
{\tt\small manu@nec-labs.com}
\and
Animesh Garg\\
Stanford University\\
{\tt\small garg@cs.stanford.edu}
\and
Silvio Savarese\\
Stanford University\\
{\tt\small ssilvio@stanford.edu}
}

\maketitle

\begin{abstract}
    
    Supervised 3D reconstruction has witnessed a significant progress through the use of deep neural networks. However, this increase in performance requires large scale annotations of 2D/3D data. In this paper, we explore inexpensive 2D supervision as an alternative for expensive 3D CAD annotation. Specifically, we use foreground masks as weak supervision through a raytrace pooling layer that enables perspective projection and backpropagation. Additionally, since the 3D reconstruction from masks is an ill posed problem, we propose to constrain the 3D reconstruction to the manifold of unlabeled realistic 3D shapes that match mask observations. We demonstrate that learning a log-barrier solution to this constrained optimization problem resembles the GAN objective, enabling the use of existing tools for training GANs. We evaluate and analyze the manifold constrained reconstruction on various datasets for single and multi-view reconstruction of both synthetic and real images.

\end{abstract}

\vspace{-15pt}
\section{Introduction} \label{sec:intro}

\input{1-intro}

\section{Prior Work}
\input{2-related_works}
\section{\mcreconfull}
\label{sec:gan}
\input{3-method}



\section{Experiments}
\label{sec:experiments}
\input{5-experiments}

\section{Conclusion}
We proposed \mcreconfull~(\mcrecon), a novel framework that makes use of foreground masks for 3D reconstruction by constraining the reconstruction to be in the space of unlabeled real 3D shapes. Additionally, we proposed a raytrace pooling layer to bridge the representation gap between 2D masks and 3D volumes.
We analyzed each component of the model through an ablation study on synthetic images. \mcrecon can successfully generate a high-quality
reconstruction from weak 2D supervision, with  reconstruction accuracy comparable to prior works that use full 3D supervision. Furthermore, we demonstrated that our model has strong generalization power for single-view real image reconstruction with noisy viewpoint estimation, hinting at better practical utility.


\paragraph{Acknowledgments}
We acknowledge the support of Nvidia and Toyota (1186781-31-UDARO) to make this work possible.

{\small
\bibliographystyle{ieee}
\bibliography{references}
}

\newpage
\setcounter{table}{0}
\renewcommand{\thesubsection}{S.\arabic{subsection}}
\renewcommand{\thetable}{S\arabic{table}}%
\setcounter{figure}{0}
\renewcommand{\thefigure}{S\arabic{figure}}%
\input{supplementary}

\end{document}

%% file: 1-intro.tex
Recovering the three-dimensional (3D) shape of an object is a fundamental attribute of human perception. This problem has been explored by a large body of work in computer vision, within domains such as structure from motion~\cite{sfm,slam} or multiview stereo~\cite{Furukawa_etal_2010,Furukawa_Ponce_2010,Goesele_etal_2010,Hernandez_Vogiatzis_2010}. While tremendous success has been achieved with conventional approaches, they often require several images to either establish accurate correspondences or ensure good coverage. This has been especially true of methods that rely on weak cues such as silhouettes~\cite{Savarese_etal_2007} or aim to recover 3D volumes rather than point clouds or surfaces~\cite{spacecarving}. In contrast, human vision seems adept at 3D shape estimation from a single or a few images, which is also a useful ability for tasks such as robotic manipulation and augmented reality.

The advent of deep neural networks has allowed incorporation of semantic concepts and prior knowledge learned from large-scale datasets of examples, which has translated into approaches that achieve 3D reconstruction from a single or sparse viewpoints~\cite{3dr2n2, singleviewproj, tlnet, 3dgan, 3dshapenet}. But conventional approaches to train convolutional neural networks (CNNs) for 3D reconstruction requires large-scale supervision. To learn the mapping from images to shapes, CAD models or point clouds are popularly used. However, ground truth alignments of models to images are challenging and expensive to acquire. Thus, existing datasets that contain an image to 3D model mappings simply label the closest model as ground truth \cite{pascal3d,objectnet3d}, which leads to suboptimal training.

\begin{figure}[t!]
    \centering
    \includegraphics[width=0.9\linewidth]{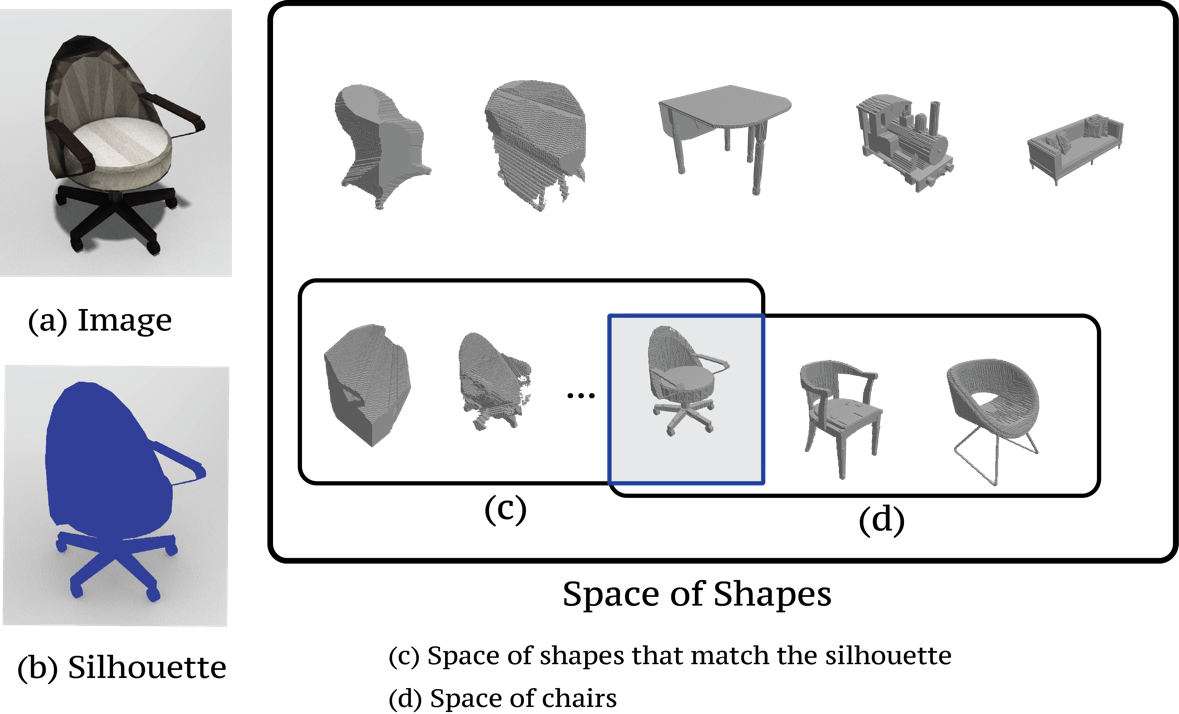}
    \caption{The 3D reconstruction using foreground masks (silhouette) is an ill-posed problem. Instead, we propose using a manifold constraint to regularize the ill posed problem.}
    \label{fig:overview}
    \vspace{-12pt}
\end{figure}

This paper presents a framework for volumetric shape reconstruction using silhouettes (foreground mask) from a single or sparse set of viewpoints and camera viewpoints as input. 
Visual hull reconstruction from such inputs is an ill-posed problem no matter how many views are given (Fig.~\ref{fig:overview}). For example, concavity cannot be recovered from silhouettes while it may contain crucial information regarding the functionality of the objects such as cups and chairs.  In addition, it is difficult to collect dense viewpoints of silhouettes of the reconstruction target in practical settings such as in online retailers. Therefore, in order solve such ill-posed problem, we regularize the space of valid solution. For example, given an image or images of a chair, we make the reconstruction to be a \textit{seatable} chair with concavity, which cannot be recovered from silhouettes. This problem becomes a constrained optimization where we solve
\begin{align}
\label{eq:constop}
\begin{split}
    \underset{x}{\text{minimize}} & \quad \text{ReprojectionError}(x) \\
    \text{subject to} & \quad \text{Reconstruction} \; x \; \text{to be a valid chair}
\end{split}
\end{align}
where $x$ is the 3D reconstruction. We denote the space of valid chairs as \textbf{the manifold of realistic shapes}, $\mathcal{M}$ which can be defined using a set of hand-designed shapes or scanned 3D shapes, denoted as $\{x^\star_i\}_i$. Then, the constraint can be written concisely as
$$\text{subject to} \; x \in \mathcal{M}$$
We solve the above constrained optimization using the log barrier method \cite{cvx} and learn the barrier function using $\{x^\star_i\}_i$. The log barrier function that we learn is similar to the discriminator in many variants of Generative Adversarial Networks \cite{manipulation,pix2pix}. We differ in framing the problem as constrained optimization to make it \textit{explicit} that we need the manifold constraint to solve such ill-posed problems and to provide a principled rationale for using an adversarial setting. Our formulation also allows clearer distinctions from other use of manifold and discriminators in Sec.~\ref{sec:log_barrier_and_gan}.

To model the reprojection error, we propose a raytrace pooling layer in Sec.~\ref{sec:raytracing} that mimics the conventional volumetric reconstruction methods such as voxel carving~\cite{spacecarving} and does not suffer from aliasing compared to \cite{singleviewproj}. Once we train the network, it only uses images at test time.

In Sec.~\ref{sec:experiments}, we experimentally evaluate our framework using three different datasets and report quantitative reductions in error compared with various baselines. Our experiments demonstrate that the proposed framework better encapsulates semantic or category-level shape information while requiring less supervision or relatively inexpensive weak supervision compared to prior works \cite{3dr2n2,singleviewproj}. In contrast to traditional voxel carving, our manifold constraint allows recovering concavities by restricting the solution to the set of plausible shapes. Quantitative advantages of our framework are established by extensive validation and ablation study on ShapeNet, ObjectNet3D and OnlineProduct datasets.



\ignore{
To summarize, the main contributions of this paper are:
\vspace{-0.2cm}
\begin{tight_itemize}
\item A novel framework based on deep neural networks to learn from weak supervision by exploiting the target manifold learned by an adversarial discriminator.
\item Application of our framework towards 3D reconstruction from a single or few segmentation masks, avoiding the requirement of expensive 3D annotation. \item Development of a perspective voxel raytracing pooling layer to link 2D annotation with 3D representation.
\item Validation on several datasets to showcase weakly supervised single view and multiview reconstruction.
\end{tight_itemize}
}

%% file: 2-related_works.tex
In this section, we briefly discuss prior works related to the three aspects of our framework: Convolutional Neural Networks for 3D data, supervised 3D reconstruction and Generative Adversarial Networks.

\vspace{3pt}
\noindent \textbf{3D Convolutional Neural Networks.}
First introduced in video classification, the 3D Convolutional Neural Networks have been widely used as a tool for spatiotemporal data analysis~\cite{3dcnn, 3dcnn2, tran15, Pan_2016, Tran_2016}. Instead of using the third dimension for temporal convolution, \cite{3dshapenet, voxnet} use the third dimension for the spatial convolution and propose 3D convolutional deep networks for 3D shape classification. Recently, 3D-CNNs have been widely used for various 3D data analysis tasks such as 3D detection or classification \cite{DeepSlidingShapes, Ren_2016_CVPR, qi2016volumetric}, semantic segmentation \cite{dai2017scannet, octnet} and  reconstruction~\cite{interpreternet, 3dr2n2, 3dgan, tlnet, singleviewproj}.
Our work is closely related to those that use the 3D-CNN for reconstruction, as discussed in the following section.


\ignore{
\vspace{3pt}
\noindent \textbf{Voxel carving.} In our work, the weak supervision using Perspective Raytrace Pooling layer in section \ref{sec:raytracing} is analogous to voxel carving~\cite{spacecarving, seitz1999photorealistic, matusik2000image} since it only requires silhouettes and camera parameters for 3D reconstruction.
Yet, \wsgan differs from voxel carving in requiring silhouette and approximate viewpoint only at training phase Further, it learns to reconstruct concavities, which are not visible from silhouettes. Finally, it can reconstruct from sparse viewpoints of masks from as little as one view. Our experiments are designed to illustrate these distinctions.
}

\vspace{3pt}
\noindent \textbf{Supervised 3D voxel reconstruction.}
Among many lines of work within the 3D reconstruction \cite{sfm, spacecarving, Furukawa_etal_2010, Furukawa_Ponce_2010, bao2013dense, dame2013dense, kar2015category, savinov2015discrete, interpreternet, rock2015completing}, ours is related to recent works that use neural networks for 3D voxel reconstruction.
\ignore{
    Many efforts has been made to reconstruct an instance by incorporating 3D geometry with
    learned priors. Bao~\etal\cite{bao2013dense} presents a dense reconstruction
    of sparse multi-view-stereo output using semantic priors.
    Dame~\etal\cite{dame2013dense} modified SLAM system to simultaneously recover
    3D shape along with its pose and scale.
    Vicente~\etal\cite{vicente2014reconstructing} aims to reconstruct a reasonable
    3D shape which best fits mask across same-class objects.
    Kar~\etal\cite{kar2015category} proposed an automated pipeline which outputs
    high-frequency 3D surfaces of known objects from 2D images.
    Rock~\etal\cite{rock2015completing} proposed a pipeline to reconstruct 3D shape
    from single depth image based on priors retrieved from a shape repository.
    Savinov~\etal\cite{savinov2015discrete} optimizes reprojection error inferring
    3D geometry with learned semantics.
    CNN based reconstruction methods largely improved the potential of data-driven approach of 3D reconstruction.
    Dosovitskiy \etal\cite{generating_chairs} first proposed a way to generate 2D
    images from a shape index and pose. The method was extended to generate 3D
    shapes using traditional multiview SfM pipeline~\cite{multiview_from_cnn}.
    Wu~\etal\cite{3dshapenet} used a Restricted Boltzmann Machine to learn the
    distribution of 3D shapes and used the RBM for 3D shape classification,
    completion, and next best view prediction.
}
Grant \etal\cite{tlnet} propose an autoencoder to learn the 3D
voxelized shape embedding and regress to the embedding from 2D images using a
CNN and generated 3D voxelized shape from a 2D image. Choy \etal\cite{3dr2n2} use a 3D-Convolutional Recurrent Neural Network to
directly reconstruct a voxelized shape from multiple images of the
object. The work of \cite{3dgan} combines a 3D-CNN with a Generative Adversarial Network to learn the latent space of 3D shapes. Given the latent space of 3D shapes, \cite{3dgan} regresses the image feature from a 2D-CNN to the latent space to reconstruct a single-view image. These approaches require associated 3D shapes for training. Recently, Yan \etal\cite{singleviewproj} propose a way to train a neural network to reconstruct 3D shapes using a large number of foreground masks (silhouettes) and viewpoints for weak supervision. The silhouette is used to carve out spaces analogous to voxel carving~\cite{spacecarving, seitz1999photorealistic, matusik2000image} and to generate the visual hull. 

Our work is different from \cite{singleviewproj, tulsiani2017multi} in that it makes use of both unmatched 3D shape and inexpensive 2D weak supervision to generate realistic 3D shapes without explicit 3D supervision. This allows the network to learn reconstruction with minimal 2D supervision (as low as one view 2D mask). And the key mechanism that allows such 2D weak supervision is the projection. Unlike \cite{singleviewproj}, we propose the Raytrace Pooling layer that is not limited to the grid sampling and experimentally compare with it in Sec.~\ref{sec:raytracing_comparison}. In addition, we use a recurrent neural network that can handle both single and multi-view images as the weak supervision is done on single or multi view images.

%% file: 3-method.tex
\begin{figure*}[ht!]
    \centering
    \includegraphics[width=0.9\linewidth]{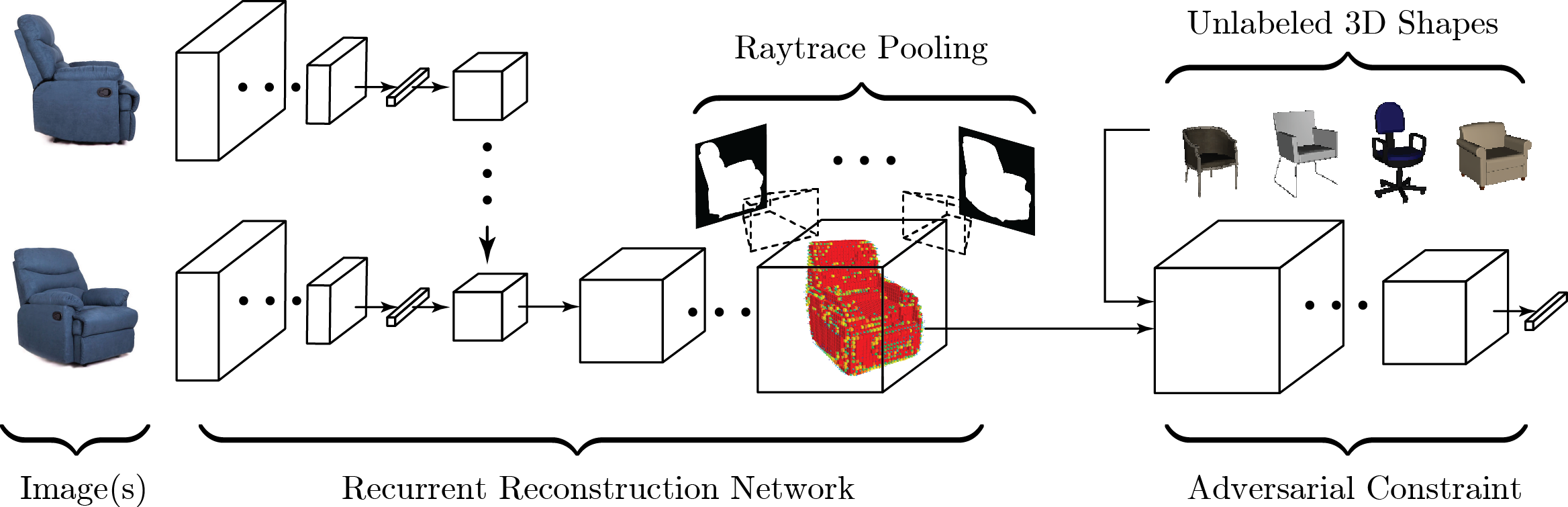}
    \caption{Visualization of \mcrecon~ network structure. Our network encodes a set of images into a latent variable. Then, the latent variable is decoded into a voxel representation of 3D shape. Perspective Raytrace Pooling layer renders this 3D shape into 2D occupancy map, allowing us to give mask supervision. Additionally, discriminator takes the generated voxel as an input, filling the missing information of the 3D shape distribution learned from unlabeled 3D models.}
    \label{fig:network}
    \vspace{-10pt}
\end{figure*}

Recent supervised single view reconstruction methods~\cite{tlnet,3dr2n2,interpreternet,3dgan} require associated 3D shapes. However, such 3D annotations are hard to acquire for real image datasets such as~\cite{imagenet,song2016deep}. Instead, we propose a framework, termed as \mcreconfull (\mcrecon), that relies on inexpensive 2D silhouette and approximate viewpoint for weak supervision. \mcrecon makes use of unlabeled 3D shapes to constrain the ill-posed single/sparse-view reconstruction problem. In this section, we propose how we solve the constrained optimization in \ref{eq:constop} using the log barrier method and show the connection between the constrained optimization and the Generative Adversarial Networks. Then we define the reprojection error using ray tracing and conclude the section with the optimization of the entire framework.


\subsection{Log Barrier for Constrained Optimization}
\label{sec:overview}

\mcrecon solves the constrained optimization problem where we minimize the reprojection error of the reconstruction while constraining the reconstruction to be in the manifold of realistic 3D shapes (Eq.~\ref{eq:constop}). Formally,
\begin{align}
\label{eq:formal_constrained_optimization}
\begin{split}
    \underset{\hat{x}}{\text{minimize}} & \quad \underset{v \in \text{views}}{\mathbb{E}} [\mathcal{L}_{\text{reproj.}}(\hat{x}, c_v, m_v)] \\
    \text{subject to} & \quad \hat{x} \in \mathcal{M}
\end{split}
\end{align}
where $L_{\text{reproj.}}(\cdot, \cdot)$ denotes the reprojection error, $x$ denotes the final reconstruction, $m_v$ and $c_v$ denote the foreground mask (silhouette) and associated camera viewpoint. We use a neural network $f(\cdot; W)$, composition of $N$ functions parametrized by $\theta_f$, to model the reconstruction function which takes multiview images $\mathbf{I}$ as an input.
\begin{align}
\hat{x} = f(\mathbf{I}; \theta_f) \quad f \coloneqq f_N \circ f_{N - 1} \circ ... \circ f_1
\end{align}
Specifically, we use the log barrier method \cite{cvx} and denote the penalty function as $g(x)$ and $g(x) = 1$ iff $x \in \mathcal{M}$ otherwise $0$. Then the constrained optimization problem in Eq.~\ref{eq:formal_constrained_optimization} becomes an unconstrained optimization problem where we solve
\begin{align}
\underset{\hat{x}}{\text{minimize}} \; \underset{v \in \text{views}}{\mathbb{E}} [\mathcal{L}_{\text{reproj.}}(\hat{x}, c_v, m_v)] - \frac{1}{t} \log g(\hat{x})
\end{align}
As $t \rightarrow \infty$, the log barrier becomes an indicator function for the constraint violation. However, the function $g(\cdot)$ involves high level cognition (does the shape look like a chair?) which captures all underlying constraints that make a 3D shape look like a valid shape: geometric constraints (symmetry, physical stability), and semantic constraints (e.g. chairs should have concavity for a seat, a backrest is next to a seat). Naturally, the function cannot be simply approximated using hand designed functions.

\subsection{Learning the Barrier for Manifold Constraint}

\begin{figure}[t]
    \centering
    \vspace{-10pt}
    \includegraphics[width=0.7\linewidth]{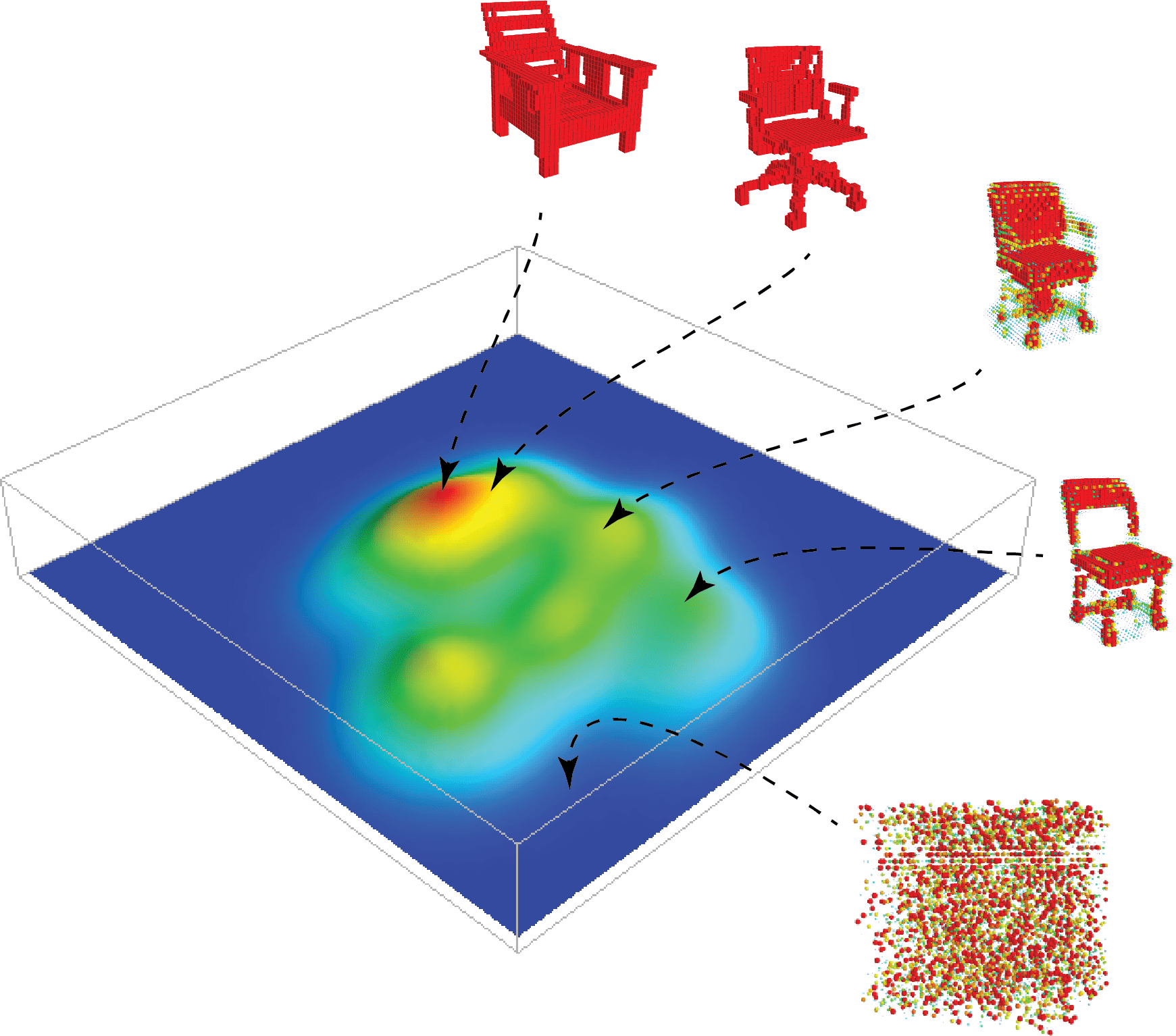}
    \caption{Illustration of the penalty function. $g(x)$ learns the manifold of realistic hand-designed or scanned 3D shapes.}
    \label{fig:gan}
    \vspace{-10pt}
\end{figure}

Instead of hand-designing the constraint violation, we learn the constraint violation function $- \log g(\cdot)$ using a neural network.
Specifically, we use the adversarial setting in \cite{gan} to \textbf{a)} adaptively learn the violation that the current generative model is violating the most, \textbf{b)} to capture constraints that are difficult to model, such as geometric constraints and semantic constraints, \textbf{c)} allow the reconstruction function to put more emphasis on the part that the current barrier focuses on as the penalty function becomes progressively more difficult.

To understand the penalty function $- \log g(\cdot)$, we should analyze the ideal scenario where the discriminator perfectly discriminates the reconstruction $\hat{x} = f(\mathbf{I})$ from the real 3D shapes $x^\star$. The ideal discriminator $g^\star(x)$ will output a value $1$ when $x$ is realistic and the log barrier will be $ - \log 1 = 0$. On the other hand, if the reconstruction is not realistic (i.e. violates any physical or semantic constraints), then the discriminator will output $0$ making the log barrier $- \log 0 = \infty$. Thus, the ideal discriminator works perfectly as the manifold constraint penalty function.

We learn the penalty function by regressing the values and minimizing the following objective function.
\begin{align}
    \underset{g}{\text{minimize}} & \quad \underset{x^\star \sim p}{\mathbb{E}} \log g(x^\star) + \underset{\hat{x} \sim q}{\mathbb{E}}\log(1-g(\hat{x}))
\end{align}
where $p$ and $q$ denote the distribution of the unlabeled 3D shapes and the reconstruction, respectively.

\subsubsection{Penalty Functions and Discriminators}
\label{sec:log_barrier_and_gan}

The log barrier we propose is similar to the discriminators in many variants of Generative Adversarial Networks that model the perceptual loss \cite{manipulation,pix2pix,shrivastava2016learning}. The discriminators work by learning the distribution of the real images and fake images and thus, it is related to learning the penalty. However, to the best of our knowledge, we are the first to make the formal connection between the discriminator and the log barrier method in constrained optimization. We provide such novel interpretation for the following reasons: \textbf{1)} to make it explicit that we need the manifold constraint to solve such ill-posed problems, \textbf{2)} to provide a principled rationale for using an adversarial network (learnable barrier) rather than simply merging the discriminator for reconstruction, \textbf{3)} to differentiate the use of the discriminator from that of \cite{3dgan} where the GAN is used ``to capture the structural difference of two 3D objects'' for feature learning, \textbf{4)} to provide a different use of manifold than that of \cite{manipulation} where manifold traversal in the latent space (noise distribution $z$) of the generators is studied. Rather, we use the manifold in the \textit{discriminator} as a barrier function.

\subsubsection{Optimal Learned Penalty Function}

However, given a fixed reconstruction function $f$, the optimum penalty function $g$ cannot discriminate a real object from the reconstruction perfectly if the distribution of the reconstruction $q(\hat{x})$ and the distribution of unlabeled hand-designed or scanned shapes $p(x^\star)$ overlap. In fact, the analysis of the optimal barrier follows that of the discriminator in \cite{gan} as the learned penalty function works and trains like a discriminator. Thus, the optimal penalty becomes $g^\star(x) = \frac{p(x)}{p(x) + q(x)}$ where $p$ is the unlabeled 3D shape. Thus, as the reconstruction function generates more realistic shapes, the constraint violation $g$ becomes less important. This behavior works in favor of the reprojection error and the reconstruction function puts more emphasis on minimizing the objective function as the reconstruction gets more realistic.

\subsection{Raytrace Pooling for Reprojection Error}
\label{sec:raytracing}
\begin{figure}[!ht]
  \centering
  \vspace{-10pt}
  \includegraphics[width=0.7\linewidth]{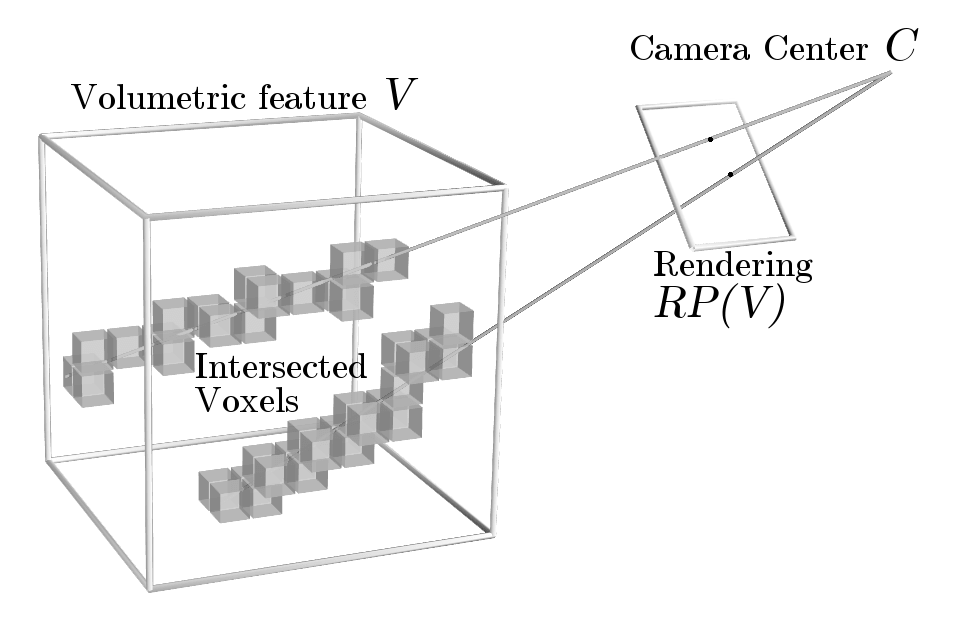}
  \caption{Visualization of raytrace pooling. For each pixel of 2D rendering,
  we calculate the direction of the ray from camera center. Then, we apply
  pooling function to all hit voxels in 3D grid.}
  \label{fig:raytracing}
  \vspace{-5pt}
\end{figure}

The 2D weak supervisions reside in the image domain whereas the reconstruction is in 3D space. To bridge different domains, we propose a Raytrace Pooling layer (\raytracinglayer).
It takes a 3D volumetric reconstruction $x$ and camera viewpoint $c$ and generates the rendering of the reconstruction $x$. Here, $c$ consists of the camera center $C$ and camera perspective $R$. Let a ray emanating from camera center $C$ be $\mathbf{L}_i$ and the intersection of the ray with the image plane be $p_i$. Then, ray can be parametrized by $u \in \mathbb{R}_+$

\vspace{-10pt}
\begin{gather}
\mathbf{L}(u) = \mathbf{C} + u\frac{\mathbf{R}^{-1}p - \mathbf{C}}{\Vert \mathbf{R}^{-1}p - \mathbf{C} \Vert}
\end{gather}
\vspace{-10pt}

We aggregate all the voxels $v_j$ that intersect with the ray $\mathbf{L}_i$ using an octree voxel-walking~\cite{arvo1988linear} with an efficient ray-box intersection algorithm~\cite{williams2005efficient}, and compute a single feature for each ray $f_i$ by pooling over the features in the voxels. We visualize the result of the raytracing and aggregated voxels in Fig.~\ref{fig:raytracing}. While multiple types of pooling operations are admissible, we use max pooling in this work. Max pooling along the ray $\mathbf{L}_i$ in an occupancy grid $x$ results in a foreground mask $\tilde{m}$.
Finally, we can measure the difference between the predicted foreground mask $\tilde{m} = RP(x, c_j)$ and the ground truth foreground mask $m$ and define a loss $\mathcal{L}_{\text{reproj.}}$:
%
\vspace{-10pt}
\begin{equation*}
\mathcal{L}_{\text{reproj.}}(x, \mathbf{c}, \mathbf{m})  = \frac{1}{M}\sum_{j}^M \mathcal{L}_{\text{s}} (RP(x, c_j), m_j),
\end{equation*}
%
where $M$ is the number of silhouettes from different viewpoints and $c_j$ is the $j$-th the camera viewpoint, and $\mathcal{L}_{\text{s}}$ is the mean of per pixel cross-entropy loss.
Instead of using raytracing for rendering, a concurrent work in \cite{singleviewproj} has independently proposed a projection layer based on the Spatial Transformer Network \cite{spatial_transformer}. Since there might be aliasing if the sampling rate is lower than the Nyquist rate~\cite{signalprocessing}, the sampling grid from \cite{singleviewproj} has to be dense and compact. To see the effect of aliasing in sampling-based projection, we compare the performance of \cite{singleviewproj} and \raytracinglayer in Sec.~\ref{sec:raytracing_comparison}. Furthermore, unlike synthetic data where the range of depth is well-controlled, depths of the target objects are unrestricted in real images, which requires dense sampling over a wide range of depth. For our real image reconstruction experiment in Sec.~\ref{sec:objectnet3d}, we determine the range of possible depths over the training data and sample over 512 steps in order to avoid the aliasing effects of \cite{singleviewproj}, far exceeding the 32 steps originally proposed there. On the other hand, \raytracinglayer mimics the rendering process and does not suffer from aliasing or depth sampling range as it is based on hit-test.

\subsection{\mcrecon Optimization}
\label{sec:optimization}


Finally, we return to the original problem of Eq.~\ref{eq:formal_constrained_optimization} and train the weakly supervised reconstruction functions given by $x = f(\mathbf{I}; \theta_f)$.
\vspace{-5pt}
\begin{align}
    \underset{\hat{x} \coloneqq f(\mathbf{I}; \theta_f)}{\text{minimize}} & \quad \underset{v \in \text{views}}{\mathbb{E}} [\mathcal{L}_{\text{reproj.}}(\hat{x}, m_v)] - \frac{1}{t} \log g(\hat{x})
\end{align}
Then we train the log barrier so that it  regresses to the ideal constraint function $g(x) = 1$ if $x \in \mathcal{M}$ and $0$ otherwise.
\begin{align}
    \underset{g}{\text{minimize}} & \quad \underset{x^\star \sim p}{\mathbb{E}} \log g(x^\star) + \underset{\hat{x} \sim q}{\mathbb{E}}\log(1-g(\hat{x}))
\end{align}
$p(x)$ is the probability distribution of the unlabeled 3D shapes and $q$ denotes the probability distribution of reconstruction $q(x|\mathbf{I})$. The final algorithm is in Algo.~\ref{alg:optimization}.

\begin{algorithm}[!h]
\caption{\mcrecon: Training}
\label{alg:optimization}
\begin{algorithmic}[1]
\Require Datasets: $\mathcal{D}_I =\{(\mathbf{I}_i, m_i, c_i)\}_i$, $\mathcal{D}_S =\{x_i^\star\}_i$
\Function{\mcrecon}{$\mathcal{D}_I, \mathcal{D}_S$}
\While{not converged}
    \ForAll{images $(\mathbf{I}_i, m_i, c_i) \in \mathcal{D}$}
        \State $\hat{x} \gets f(\mathbf{I}_i)$ \Comment{3D reconstruction}
        \ForAll{camera $c_{i,j}$, s.t. $j \in \{1,\cdots,M\}$}
            \State $\tilde{m}_{i,j} \gets RP(x, c_{i,j})$ \Comment{Reprojection}
        \EndFor
        \State $g \gets \text{UpdatePenalty}(\hat{x}, x^\star)$
        \State $\mathbb{E} [\mathcal{L}_{\text{reproj.}}] \gets \frac{1}{M}\sum_{j = 1}^M \mathcal{L}_s(\tilde{m}_{i,j}, m_{i,j})$
        \State $\mathcal{L}_f \gets \mathbb{E}[\mathcal{L}_{reproj.}] - \frac{1}{t}\log g(x)$
        \State $\theta_f \gets \theta_f - \alpha \partial \mathcal{L}_f / \partial \theta_f$    \EndFor
\EndWhile
\State \Return $f$
\EndFunction
\end{algorithmic}
\end{algorithm}

\begin{algorithm}[!h]
\caption{Penalty Function Update}
\label{alg:penalty}
\begin{algorithmic}[1]
\Require Datasets: reconstruction $\hat{x}$ and unlabeled 3D shapes $x^\star$
\Function{UpdatePenalty}{$\hat{x}, x^\star$}
\State \begin{varwidth}[t]{\linewidth}
    $L_g \gets \frac{1}{|\hat{x}|}\sum_{i \in |\hat{x}|} \log g(\hat{x}_i)$ \par
    \hskip \algorithmicindent $ + \frac{1}{|x^\star|} \sum_{i \in |x^\star|} \log(1-g(x_i^\star))$
\end{varwidth}
\State $\theta_g \gets \theta_g - \alpha \partial \mathcal{L}_g / \partial \theta_g$
\State \Return $g$
\EndFunction
\end{algorithmic}
\end{algorithm}

While convergence properties of such an optimization problem are nontrivial to prove and an active area of research, our empirical results consistently indicate it behaves reasonably well in practice.

%% file: 5-experiments.tex
To validate our approach, we design various experiments and use standard datasets.
First, we define the baseline methods including recent works (Sec.~\ref{sec:baselines}) and evaluation metrics (Sec.~\ref{sec:evaluation}).
To compare our approach with baseline methods in a controlled environment, we use a 3D shape dataset and rendering images. We present quantitative ablation study results on Sec.~\ref{sec:shapenet}. Next, we test our framework on a real image single-view and a multi-view dataset in Sec.~\ref{sec:objectnet3d} and Sec.~\ref{sec:stanfordonline} respectively. To examine the expressive power of the reconstruction function $f$, we examine the intermediate representation and analyze its semantic content in Sec~\ref{sec:anaylsis} similar to \cite{dcgan, 3dgan}. Note that, we can manipulate the output (shape) using a different modality (image) and allow editing in a different domain.

\begin{table*}[ht!]
    \scriptsize
    \centering
    \begin{tabular}{|c|c|c|c|c|c|c|c|c|}
    \multicolumn{9}{c}{{\small IOU / AP}}\\
        \hline \multirow{2}{*}{Level of supervision}&\multirow{2}{*}{Methods}&\multicolumn{2}{c|}{Transportation}&\multicolumn{4}{c|}{Furniture}&\multirow{2}{*}{Mean}\\ \cline{3-8}
               &                         & car            & airplane & sofa   & chair  & table     & bench &    \\ \hline
\multirow{3}{*}{1 view 2D}               & VC~\cite{spacecarving} & 0.2605 / 0.2402         & 0.1092 / 0.0806   & 0.2627 / 0.2451 & 0.2035 / 0.1852 & 0.1735 / 0.1546    & 0.1303 / 0.1064 & 0.1986 / 0.1781 \\
     & PTN~\cite{singleviewproj} & 0.4437 / 0.7725         & 0.3352 / 0.5568   & 0.3309 / 0.4947 & 0.2241 / 0.3178 & 0.1977 / 0.2800    & 0.2145 / 0.2884 & 0.2931 / 0.4620 \\
               & RP                     & 0.3791 / 0.7250         & 0.2508 / 0.4997   & 0.3427 / 0.5093 & 0.1930 / 0.3361  & 0.1821 / 0.2664    & 0.2188 / 0.3003 & 0.2577 / 0.4452 \\ \hline
\multirow{2}{*}{1 view 2D + U3D}          & RP+NN                  & 0.5451 / 0.5582         & 0.2057 / 0.1560   & 0.2767 / 0.2285 & 0.1556 / 0.1056 & 0.1285 / 0.0872    & 0.1758 / 0.1183 & 0.2597 / 0.2267 \\
               &  \mcrecon                  & \textbf{0.5622} / \textbf{0.8244}         & \textbf{0.3727} / \textbf{0.5911}   & \textbf{0.3791} / \textbf{0.5597} & \textbf{0.3503} / \textbf{0.4828} & \textbf{0.3532} / \textbf{0.4582}    & \textbf{0.2953} / \textbf{0.3912} & \textbf{0.4036} / \textbf{0.5729} \\ \hline \hline
\multirow{3}{*}{5 views 2D}              & VC~\cite{spacecarving} & 0.5784 / 0.5430         & 0.3452 / 0.2936   & 0.5257 / 0.4941 & 0.4048 / 0.3509 & 0.3549 / 0.3011    & 0.3387 / 0.2788 & 0.4336 / 0.3857 \\
      & PTN~\cite{singleviewproj} & 0.6593 / 0.8504         & 0.4422 / 0.6721   & 0.5188 / 0.7180 & 0.3736 / 0.5081 & 0.3556 / 0.5367    & 0.3374 / 0.4725 & 0.4572 / 0.6409 \\
               & RP                     & 0.6521 / \textbf{0.8713}        & 0.4344 / 0.6694  & 0.5242 / 0.7023 & 0.3717 / 0.5048 & 0.3197 / 0.4464    & 0.321 / 0.4377  & 0.4442 / 0.6123 \\ \hline
\multirow{2}{*}{5 views 2D + U3D} & RP+NN                  & \textbf{0.6744} / 0.6508        & \textbf{0.4671} / 0.4187   & 0.5467 / 0.5079 & 0.3449 / 0.2829 & 0.3081 / 0.2501    & 0.3116 / 0.2477 & 0.4465 / 0.3985 \\
               & \mcrecon                   & 0.6142 / 0.8674        & 0.4523 / \textbf{0.6877} & 0.5458 / \textbf{0.7473} & \textbf{0.4365} / \textbf{0.6212} & \textbf{0.4204} / \textbf{0.5741}    & \textbf{0.4009} / \textbf{0.5770} & \textbf{0.4849} / \textbf{0.6851} \\ \hline \hline
F3D        & R2N2~\cite{3dr2n2}                 & 0.8338 / 0.9631         & 0.5425 / 0.7747    & 0.6784 / 0.8582  & 0.5174 / 0.7266  & 0.5589 / 0.7754     & 0.4950 / 0.6982  & 0.6210 / 0.8123 \\\hline
    \end{tabular}
\vspace{0.1cm}
    \caption{Per-category 3D reconstruction Intersection-over-Union(IOU) / Average Precision(AP). Please see Sec.~\ref{sec:baselines} for details of baseline methods and the level of supervision. \mcrecon outperforms other baselines by larger margin in classes with more complicated shapes as shown in Fig. \ref{fig:shapenet}.}
    \label{tab:shapenet-ap}
\vspace{-0.2cm}
\end{table*}

\newcommand{\ha}[0]{0.07}
\newcommand{\wa}[0]{0.07}
\begin{figure*}[ht!]
    \centering
    \begin{tabular}{ccccccccc}
    \toprule
    Input &
    \includegraphics[height=\ha\linewidth,width=\wa\linewidth,keepaspectratio]{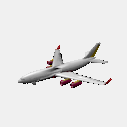} &
    \includegraphics[height=\ha\linewidth,width=\wa\linewidth,keepaspectratio]{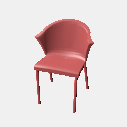} &
    \includegraphics[height=\ha\linewidth,width=\wa\linewidth,keepaspectratio]{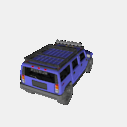} &
    \includegraphics[height=\ha\linewidth,width=\wa\linewidth,keepaspectratio]{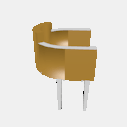} &
    \includegraphics[height=\ha\linewidth,width=\wa\linewidth,keepaspectratio]{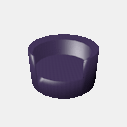} &
    \includegraphics[height=\ha\linewidth,width=\wa\linewidth,keepaspectratio]{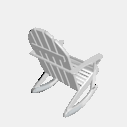} &
    \includegraphics[height=\ha\linewidth,width=\wa\linewidth,keepaspectratio]{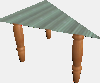} &
    \includegraphics[height=\ha\linewidth,width=\wa\linewidth,keepaspectratio]{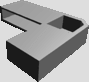} \\
    \midrule
    \specialcell{G.T.}&
    \includegraphics[height=\ha\linewidth,width=\wa\linewidth,keepaspectratio]{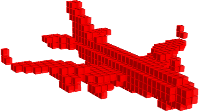} &
    \includegraphics[height=\ha\linewidth,width=\wa\linewidth,keepaspectratio]{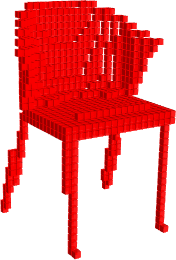} &
    \includegraphics[height=\ha\linewidth,width=\wa\linewidth,keepaspectratio]{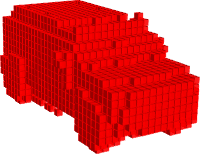} &
    \includegraphics[height=\ha\linewidth,width=\wa\linewidth,keepaspectratio]{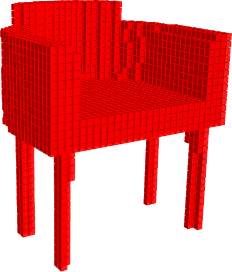} &
    \includegraphics[height=\ha\linewidth,width=\wa\linewidth,keepaspectratio]{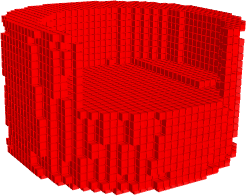} &
    \includegraphics[height=\ha\linewidth,width=\wa\linewidth,keepaspectratio]{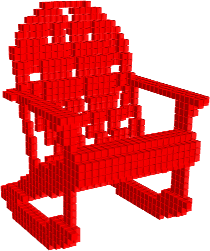} &
    \includegraphics[height=\ha\linewidth,width=\wa\linewidth,keepaspectratio]{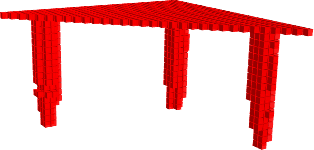} &
    \includegraphics[height=\ha\linewidth,width=\wa\linewidth,keepaspectratio]{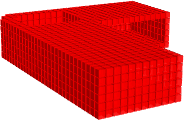} \\
    \midrule
    \specialcell{1 view\\RP}&
    \includegraphics[height=\ha\linewidth,width=\wa\linewidth,keepaspectratio]{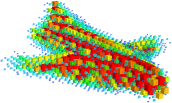} &
    \includegraphics[height=\ha\linewidth,width=\wa\linewidth,keepaspectratio]{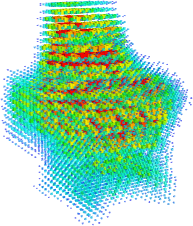} &
    \includegraphics[height=\ha\linewidth,width=\wa\linewidth,keepaspectratio]{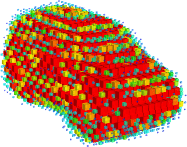} &
    \includegraphics[height=\ha\linewidth,width=\wa\linewidth,keepaspectratio]{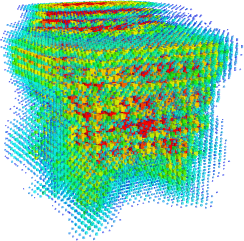} &
    \includegraphics[height=\ha\linewidth,width=\wa\linewidth,keepaspectratio]{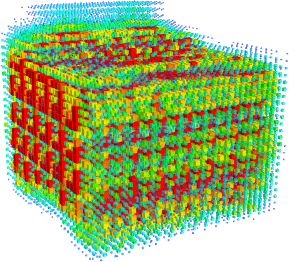} &
    \includegraphics[height=\ha\linewidth,width=\wa\linewidth,keepaspectratio]{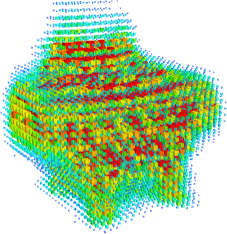} &
    \includegraphics[height=\ha\linewidth,width=\wa\linewidth,keepaspectratio]{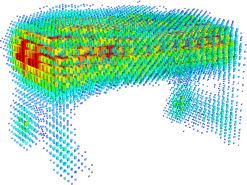} &
    \includegraphics[height=\ha\linewidth,width=\wa\linewidth,keepaspectratio]{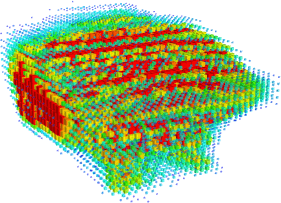} \\
    \specialcell{1 view\\\mcrecon} &
    \includegraphics[height=\ha\linewidth,width=\wa\linewidth,keepaspectratio]{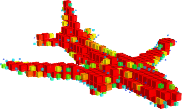} &
    \includegraphics[height=\ha\linewidth,width=\wa\linewidth,keepaspectratio]{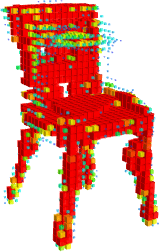} &
    \includegraphics[height=\ha\linewidth,width=\wa\linewidth,keepaspectratio]{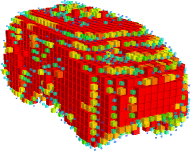} &
    \includegraphics[height=\ha\linewidth,width=\wa\linewidth,keepaspectratio]{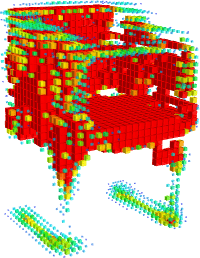} &
    \includegraphics[height=\ha\linewidth,width=\wa\linewidth,keepaspectratio]{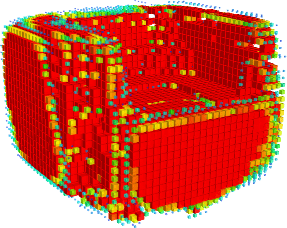} &
    \includegraphics[height=\ha\linewidth,width=\wa\linewidth,keepaspectratio]{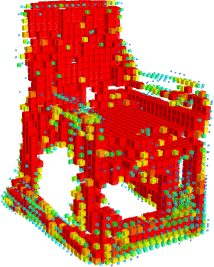} &
    \includegraphics[height=\ha\linewidth,width=\wa\linewidth,keepaspectratio]{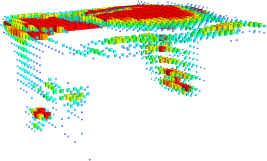} &
    \includegraphics[height=\ha\linewidth,width=\wa\linewidth,keepaspectratio]{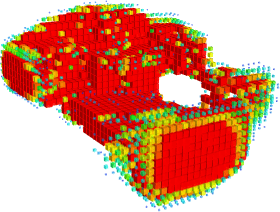} \\
    \midrule
    \specialcell{5 views\\RP}&
    \includegraphics[height=\ha\linewidth,width=\wa\linewidth,keepaspectratio]{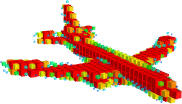} &
    \includegraphics[height=\ha\linewidth,width=\wa\linewidth,keepaspectratio]{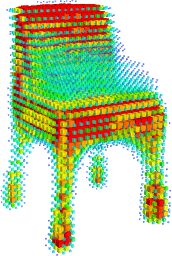} &
    \includegraphics[height=\ha\linewidth,width=\wa\linewidth,keepaspectratio]{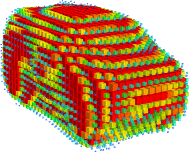} &
    \includegraphics[height=\ha\linewidth,width=\wa\linewidth,keepaspectratio]{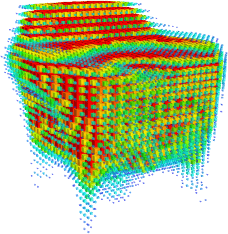} &
    \includegraphics[height=\ha\linewidth,width=\wa\linewidth,keepaspectratio]{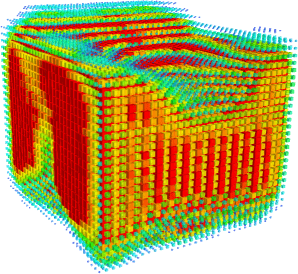} &
    \includegraphics[height=\ha\linewidth,width=\wa\linewidth,keepaspectratio]{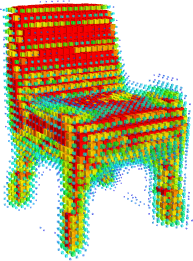} &
    \includegraphics[height=\ha\linewidth,width=\wa\linewidth,keepaspectratio]{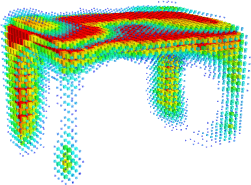} &
    \includegraphics[height=\ha\linewidth,width=\wa\linewidth,keepaspectratio]{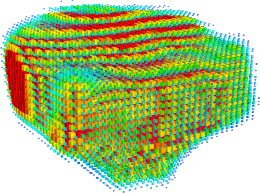} \\
    \specialcell{5 views\\\mcrecon}&
    \includegraphics[height=\ha\linewidth,width=\wa\linewidth,keepaspectratio]{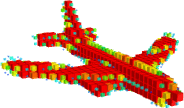} &
    \includegraphics[height=\ha\linewidth,width=\wa\linewidth,keepaspectratio]{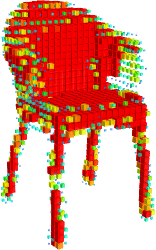} &
    \includegraphics[height=\ha\linewidth,width=\wa\linewidth,keepaspectratio]{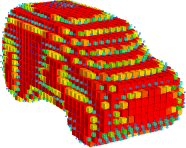} &
    \includegraphics[height=\ha\linewidth,width=\wa\linewidth,keepaspectratio]{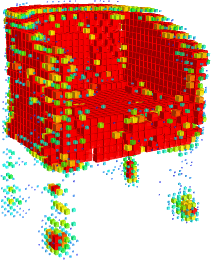} &
    \includegraphics[height=\ha\linewidth,width=\wa\linewidth,keepaspectratio]{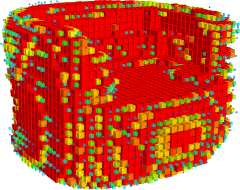} &
    \includegraphics[height=\ha\linewidth,width=\wa\linewidth,keepaspectratio]{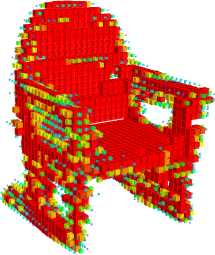} &
    \includegraphics[height=\ha\linewidth,width=\wa\linewidth,keepaspectratio]{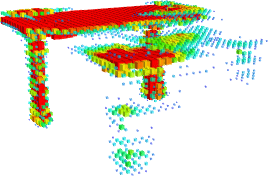} &
    \includegraphics[height=\ha\linewidth,width=\wa\linewidth,keepaspectratio]{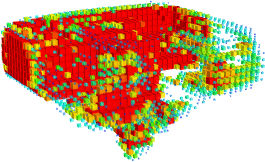} \\
    \bottomrule
    \end{tabular}
\vspace{0.1cm}
    \caption{Qualitative results of single- or multi-view synthetic image reconstructions on ShapeNet dataset. Compared to RP which only uses 2D weak supervision, \mcrecon reconstructs complex shapes better. Please refer to Sec. \ref{sec:evaluation} for details of our visualization method.}
    \label{fig:shapenet}
\vspace{-0.3cm}
\end{figure*}

\subsection{Baselines}
\label{sec:baselines}

For an accurate ablation study, we propose various baselines to examine each component in isolation. First, we categorize all the baseline methods into three categories based on the level of supervision: \textit{2D Weak Supervision (2D)}, \textit{2D Weak Supervision + unlabeled 3D Supervision (2D + U3D)}, and \textit{Full 3D Supervision (F3D)}. \textit{2D} has access to 2D silhouettes and viewpoints as supervision; and \textit{2D + U3D} uses silhouettes, viewpoints, and unlabeled 3D shapes for supervision. Finally, \textit{F3D} is supervised with the ground truth 3D reconstruction associated with the images. Given \textit{F3D} supervision, silhouettes do not add any information, thus the performance of a system with full supervision provides an approximate performance upper bound.

Specifically, in the \textit{2D} case, we use Raytrace Pooling (RP) as proposed in Sec~.\ref{sec:raytracing} and compare it with Perspective Transformer (PTN) by Yan \etal~\cite{singleviewproj}. Next, in the \textit{2D + U3D} case, we use RP + Nearest Neighbor (RP+NN) and \mcrecon. RP + NN uses unlabeled 3D shapes, by retrieving the 3D shape that is closest to the prediction. Finally, in the \textit{F3D} case, we use R2N2~\cite{3dr2n2}. We did not include \cite{3dgan, tlnet} in this experiment since they are restricted to single-view reconstruction and use full 3D supervision which would only provide an additional upper bound. For all neural network based baselines, we used the same base network architecture (encoder and generator) to ascribe performance gain only to the supervision mode. Aside from learning-based methods, we also provide a lower-bound on performance using voxel carving (VC)~\cite{spacecarving}. We note that voxel carving requires silhouette and camera viewpoint during testing. 
Kindly refer to the supplementary material for details of baseline methods, implementation, and training.

\subsection{Metrics and Visualization}
\label{sec:evaluation}

The network generates a voxelized reconstruction, and for each voxel, we have occupancy probability (confidence). We use Average Precision (AP) to evaluate the quality and the confidence of the reconstruction. We also binarize the probability and report Intersection-over-Union (IOU) with threshold 0.4, following \cite{3dr2n2}. This metric gives more accurate evaluation of deterministic methods like voxel carving.
For visualization, we use red to indicate voxels with occupancy probability above 0.6 and gradually make it smaller and green until occupancy probability reaches 0.1. When the probability is below 0.1, we did not visualize the voxel.


\subsection{Ablation Study on ShapeNet \cite{shapenet}}
\label{sec:shapenet}

\newcommand{\hc}[0]{0.14}
\newcommand{\wc}[0]{0.14}
\begin{figure}[ht!]
    \small
    \centering
    \setlength\extrarowheight{2pt}
    \begin{tabular}{ccccc}
    \toprule
    Input &
    \includegraphics[height=\hc\linewidth,width=\wc\linewidth,keepaspectratio]{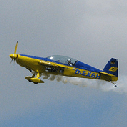} &
    \includegraphics[height=\hc\linewidth,width=\wc\linewidth,keepaspectratio]{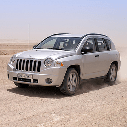} &
    \includegraphics[height=\hc\linewidth,width=\wc\linewidth,keepaspectratio]{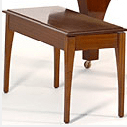} &
    \includegraphics[height=\hc\linewidth,width=\wc\linewidth,keepaspectratio]{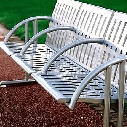} \\
    \midrule
    \specialcell{G.T.}&
    \includegraphics[height=\hc\linewidth,width=\wc\linewidth,keepaspectratio]{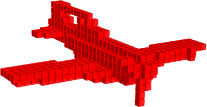} &
    \includegraphics[height=\hc\linewidth,width=\wc\linewidth,keepaspectratio]{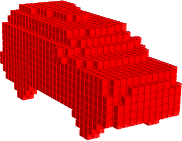} &
    \includegraphics[height=\hc\linewidth,width=\wc\linewidth,keepaspectratio]{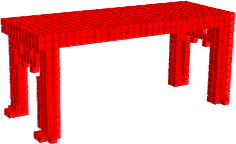} &
    \includegraphics[height=\hc\linewidth,width=\wc\linewidth,keepaspectratio]{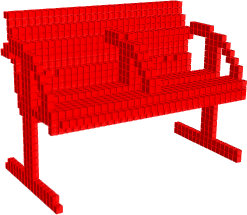} \\
    \midrule
    \specialcell{VC}&
    \includegraphics[height=\hc\linewidth,width=\wc\linewidth,keepaspectratio]{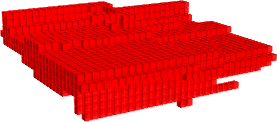} &
    \includegraphics[height=\hc\linewidth,width=\wc\linewidth,keepaspectratio]{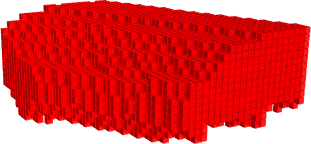} &
    \includegraphics[height=\hc\linewidth,width=\wc\linewidth,keepaspectratio]{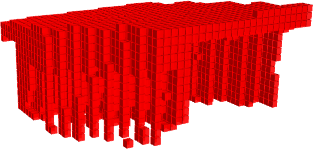} &
    \includegraphics[height=\hc\linewidth,width=\wc\linewidth,keepaspectratio]{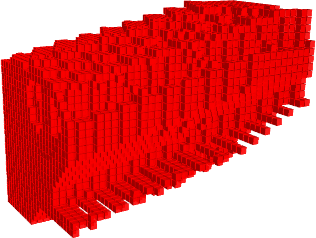} \\
    \mcrecon &
    \includegraphics[height=\hc\linewidth,width=\wc\linewidth,keepaspectratio]{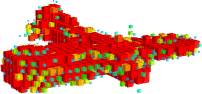} &
    \includegraphics[height=\hc\linewidth,width=\wc\linewidth,keepaspectratio]{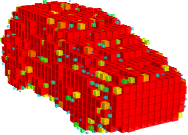} &
    \includegraphics[height=\hc\linewidth,width=\wc\linewidth,keepaspectratio]{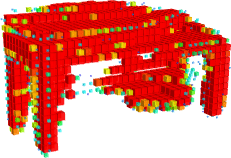} &
    \includegraphics[height=\hc\linewidth,width=\wc\linewidth,keepaspectratio]{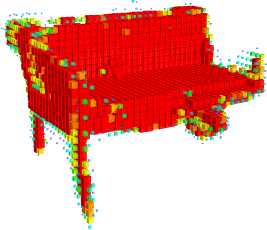} \\
    \bottomrule
    \end{tabular}
\vspace{0.3cm}
    \caption{Real image single-view reconstructions on ObjectNet3D. Compared to RP which only uses 2D weak supervision, \mcrecon reconstructs complex shapes better. Please refer to Sec. \ref{sec:evaluation} for details of our visualization method.}
    \label{fig:objectnet3d}
\end{figure}

\begin{figure}[ht!]
    \centering
    \begin{tabular}{ccccc}
    \toprule
    Input &
    \includegraphics[height=\hc\linewidth,width=\wc\linewidth,keepaspectratio]{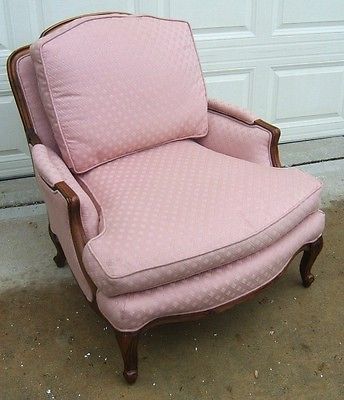} &
    \includegraphics[height=\hc\linewidth,width=\wc\linewidth,keepaspectratio]{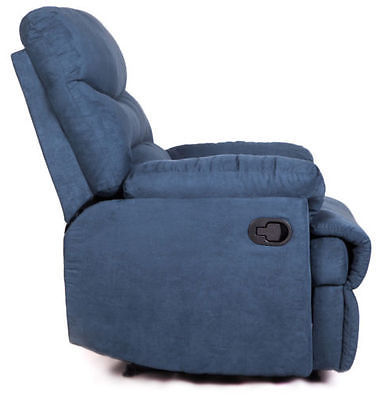} &
    \includegraphics[height=\hc\linewidth,width=\wc\linewidth,keepaspectratio]{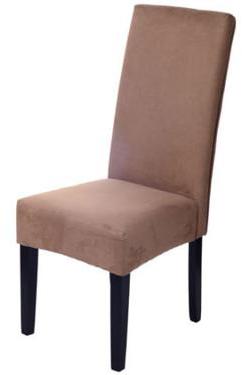} &
    \includegraphics[height=\hc\linewidth,width=\wc\linewidth,keepaspectratio]{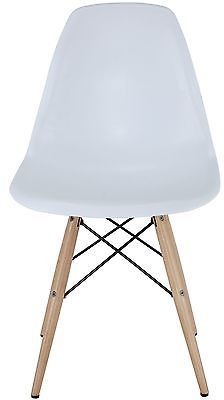} \\
    &
    \includegraphics[height=\hc\linewidth,width=\wc\linewidth,keepaspectratio]{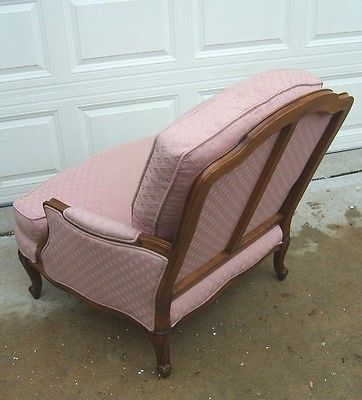} &
    \includegraphics[height=\hc\linewidth,width=\wc\linewidth,keepaspectratio]{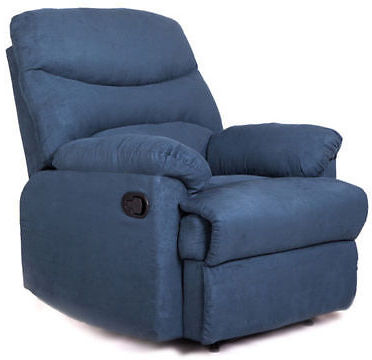} &
    \includegraphics[height=\hc\linewidth,width=\wc\linewidth,keepaspectratio]{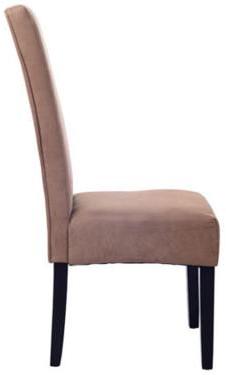} &
    \includegraphics[height=\hc\linewidth,width=\wc\linewidth,keepaspectratio]{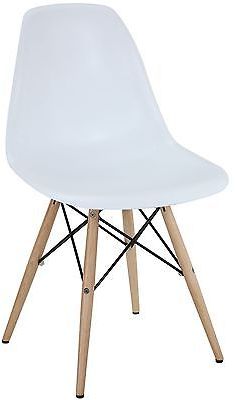} \\
    \midrule
    \mcrecon &
    \includegraphics[height=\hc\linewidth,width=\wc\linewidth,keepaspectratio]{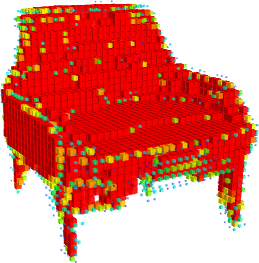} &
    \includegraphics[height=\hc\linewidth,width=\wc\linewidth,keepaspectratio]{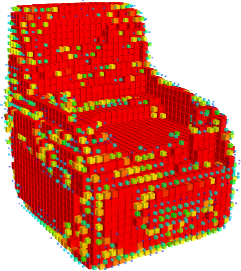} &
    \includegraphics[height=\hc\linewidth,width=\wc\linewidth,keepaspectratio]{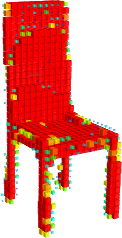} &
    \includegraphics[height=\hc\linewidth,width=\wc\linewidth,keepaspectratio]{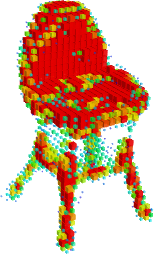} \\
    \bottomrule
    \end{tabular}
    \vspace{0.3cm}
    \caption{Qualitative results of multi-view real image reconstructions on Stanford Online Product dataset~\cite{song2016deep}. Our network successfully reconstructed real images coordinating different views.}
    \label{fig:stanfordproduct}
    \vspace{-0.5cm}
\end{figure}

\begin{figure}[h]
    \centering
    \begin{tabular}{c}
        \includegraphics[width=0.4\textwidth]{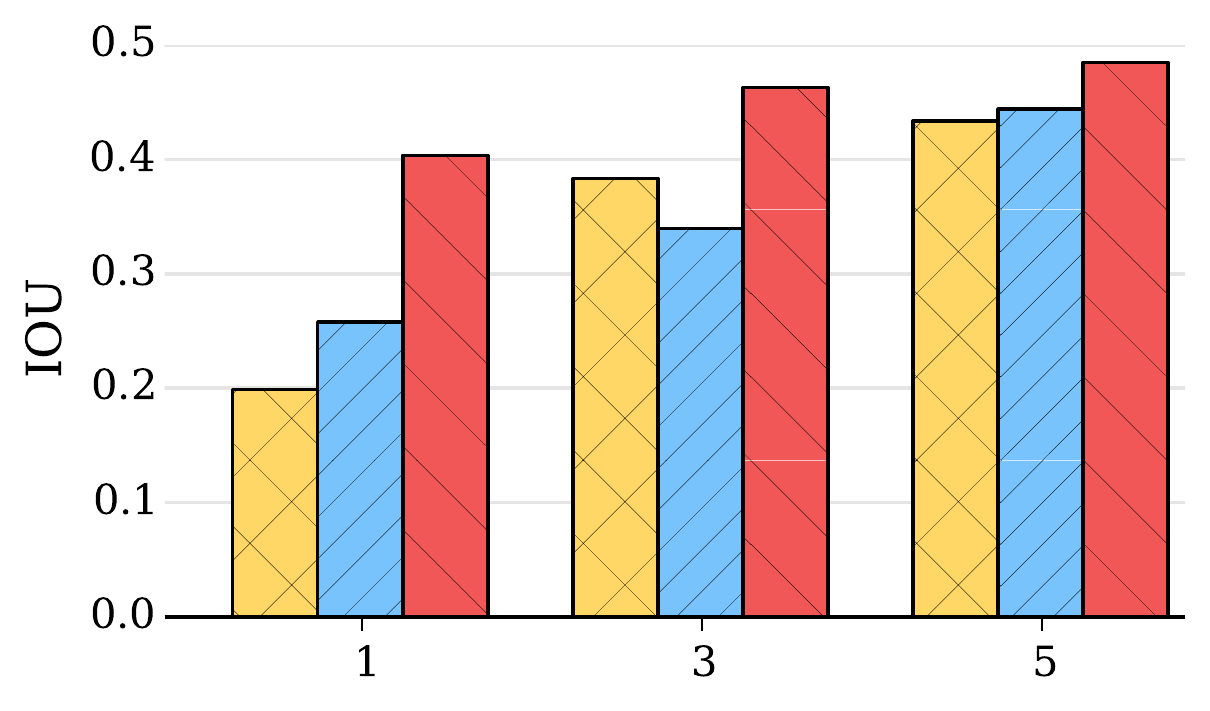} \\ \vspace{-0.2cm}
        \includegraphics[width=0.4\textwidth]{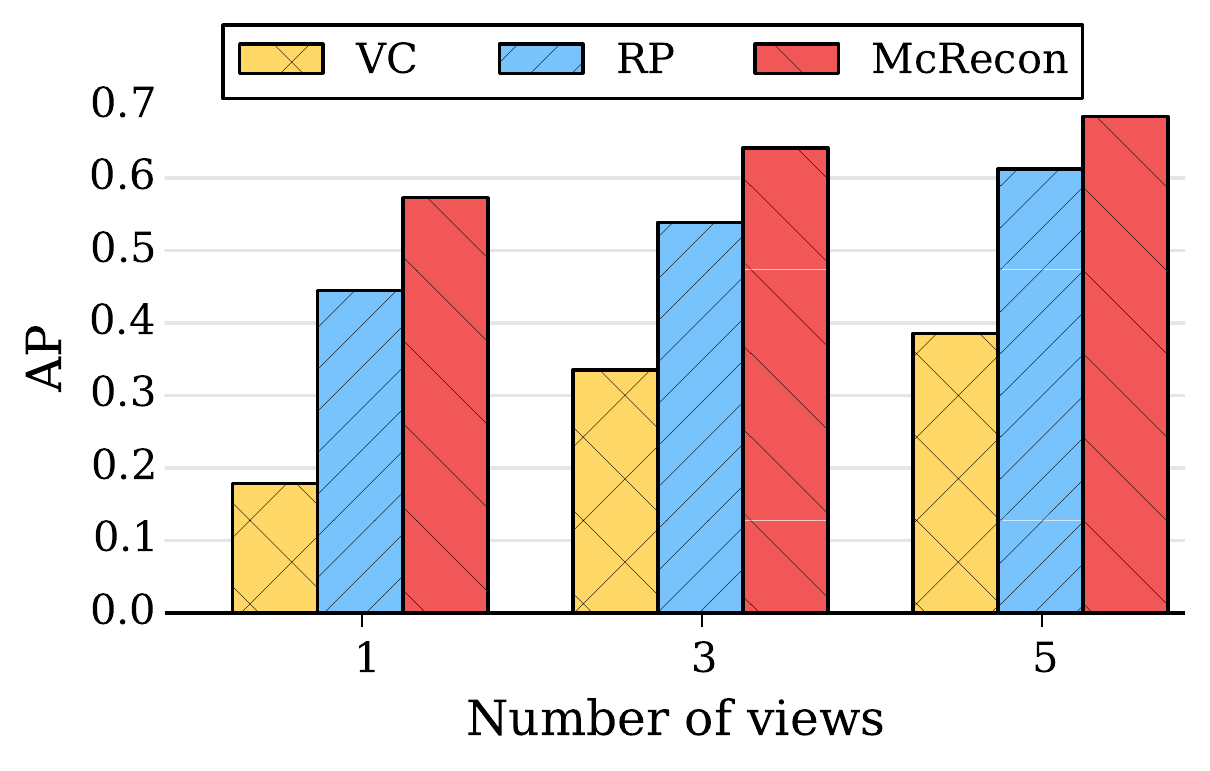}
    \end{tabular}
    \caption{Intersection-over-union (IOU) and Average Precision (AP) over the number of masks used for weak supervision. The performance gap between \mcrecon and the other baselines gets larger as the number of views of masks decreases (i.e. supervision strength gets weaker).}
    \label{fig:shapenet-plots}
    \vspace{-0.5cm}
\end{figure}

In this section, we perform ablation study and compare \mcrecon with the baseline methods on the ShapeNet~\cite{shapenet} dataset. The synthetic dataset allows us to control external factors such as the number of viewpoints, quality of mask and is ideal for ablation study. Specifically, we use the renderings from \cite{3dr2n2} since it contains a large number of images from various viewpoints and the camera model has more degree of freedom. In order to train the network on multiple categories while maintaining a semantically meaningful manifold across different classes, we divide the categories into furniture (sofa, chair, bench, table) and vehicles (car, airplane) classes and trained networks separately. We use the alpha channel of the renderings image to generate 2D mask supervisions (finite depth to indicate foreground silhouette). For the unlabeled 3D shapes, we simply voxelized the 3D shapes. To simulate realistic scenario, we divide the dataset into three \textbf{disjoint} sets: shapes for 2D weak supervision, shapes for unlabeled 3D shapes, and the test set. Next, we study the impact of the level of supervision, the number of viewpoints, and the object category on the performance.

First, we found that more supervision leads to better reconstruction and \mcrecon make use of the unlabeled 3D shapes effectively (Vertical axis of Tab.~\ref{tab:shapenet-ap}). Compare with the simple nearest neighbor, which also make use of the unlabeled 3D data, \mcrecon outperforms the simple baseline by a large margin. This hints that the barrier function smoothly interpolates the manifold of 3D shapes and provide strong guidance.
Second, \mcrecon learns to generate better reconstruction even from a small number of 2D weak supervision. In Tab.~\ref{tab:shapenet-ap} and in Fig.~\ref{fig:shapenet-plots}, we vary the number of 2D silhouettes that we used to train the networks and observe that the performance improvement that we get from exploiting the unlabeled 3D shapes gets larger as we use a fewer number of 2D supervision.
Third, we observed that \mcrecon outperforms other baselines by a larger margin on classes with more complicated shapes such as chair, bench, and table which have concavity that is difficult to recover only using 2D silhouettes. For categories with simpler shapes such as car,
the marginal benefit of using the adversarial network is smaller. Similarly, 3D nearest neighbor retrieval improves reconstruction quality only on few categories of a simple shape such as car while it also harms the reconstruction on complex shapes such as chair or table. This is
expected since their 3D shapes are close to convex shapes and 2D supervision is enough to recover 3D shapes.


\begin{table}[t]
    \vspace{0.3cm}
    \scriptsize
    \centering
    \begin{tabular}{|l|llll|l|}\hline
    &sofa     & chair  & table  & bench  & mean     \\ \hline
    PTN\_16~\cite{singleviewproj}  & 0.4753 & 0.2888 & 0.2476 & 0.2576 & 0.2979 \\
    PTN\_32~\cite{singleviewproj}  & 0.4947 & 0.3178 & 0.2800 & 0.2884 & 0.3283 \\
    PTN\_64~\cite{singleviewproj}  & 0.5082 & 0.3377 & 0.3114 & 0.3104 & 0.3509 \\
    PTN\_128~\cite{singleviewproj} & 0.5217 & 0.3424 & 0.3104 & 0.3146 & 0.3545\\ \hline
    RP   & 0.5093 & 0.3361 & 0.2664 & 0.3003 & 0.3308 \\ \hline
    \end{tabular}
    \vspace{0.3cm}
    \caption{AP of 2D weak supervision methods on single-view furniture reconstruction. In order to analyze the effect of aliasing of PTN~\cite{singleviewproj}, we varied its disparity sampling density (sampling density $N$, for all PTN\_$N$) and compare with RP.}
    \label{tab:raytracing_result}
    \vspace{-0.5cm}
\end{table}

We visualize the reconstructions in Fig.~\ref{fig:shapenet}. We observe that our network can carve concavities, which is difficult to learn solely from mask supervision and demonstrates a qualitative benefit of our manifold constraint. Also, compared to the network trained only using mask supervision, \mcrecon prefers to binarize the occupancy probability, which seems to be an artifact of the generator fooling the discriminator.


\paragraph{Raytracing Comparison}
\label{sec:raytracing_comparison}
In this section, we compare a raytracing based projection (\raytracinglayer) and a sampling based projection (PTN~\cite{singleviewproj}) experimentally on ShapeNet single view furniture category. We only vary the projection method and sampling rate along depth but keep the same base network architecture. As shown in Table. \ref{tab:raytracing_result}, the reconstruction performance improves as the sampling rate increases as expected in Sec.~\ref{sec:raytracing}. We suspect that the trilinear interpolation in PTN played a significant role after it reaches resolution 64 and that implementing a similar scheme using ray length in \raytracinglayer could potentially improve the result.

\subsection{Single-view reconst. on ObjectNet3D \cite{objectnet3d}}
\label{sec:objectnet3d}


In this experiment, we train our network for single real-image reconstruction using the  ObjectNet3D~\cite{objectnet3d} dataset. The dataset contains 3D annotations in the form of the closest 3D shape from ShapeNet and viewpoint alignment. Thus, we generate 2D silhouettes using 3D shapes. We split the dataset using the shape index to generate disjoint sets like the previous experiments. Since the dataset consists of at most 1,000 instances per category, we freeze the generator and discriminator and fine-tune only the 2D encoder $E(u)$. We quantitatively evaluate  intersection-over-union(IOU) on the reconstruction results as shown in Table \ref{tab:objectnet3d}. The numbers indicate that \mcrecon~has better generalization power beyond the issue of ill-conditioned visual hull reconstruction and silhouette-based learning~\cite{singleviewproj} from a single-view mask. Please note that voxel carving, unlike \mcrecon, requires camera parameters at test time. Qualitative results are presented in Fig.~\ref{fig:objectnet3d}.

\begin{table}[ht]
    \scriptsize
    \centering
    \begin{tabular}{| c || c | c | c | c | c | c | c |} \hline
        & sofa & chair & bench & car & airplane \\ \hline
        VC~\cite{spacecarving} & 0.304 & 0.177 & 0.146 & 0.481 & 0.151 \\ \hline
        PTN~\cite{singleviewproj} & 0.276 & 0.151 & 0.095 & 0.421 & 0.130 \\ \hline
        \mcrecon & \textbf{0.423} & \textbf{0.380} & \textbf{0.380} & \textbf{0.649} & \textbf{0.322} \\ \hline \hline
        PTN-NV~\cite{singleviewproj} & 0.207 & 0.128 & 0.068 & 0.344 & 0.100 \\ \hline
        \mcrecon-NV & \textbf{0.256} & \textbf{0.157} & \textbf{0.086} & \textbf{0.488} & \textbf{0.214} \\ \hline
    \end{tabular}
\vspace{0.3cm}
    \caption{Per-class real image 3D reconstruction intersection-over-union(IOU) percentage on ObjectNet3D. NV denotes a network trained with noisy viewpoint estimation.}
    \label{tab:objectnet3d}
    \vspace{-0.5cm}
\end{table}

\paragraph{Training with noisy viewpoint estimation}
\label{sec:viewpoint_estimation}
In this experiment, we do a noisy estimation of camera parameters instead of using the ground-truth label as an input to RP, training the network only using 2D silhouette. We estimate camera parameters by discretizing azimuth, elevation, and depth of the camera into 10 bins and finding the combination of parameters that minimize the $L_2$ distance of the rendering of a roughly aligned 3D model~\cite{objectnet3d} with the ground-truth 2D silhouette. We quantitatively evaluate intersection-over-union(IOU) on the reconstruction results as shown in Table \ref{tab:objectnet3d}. These results demonstrate that \mcrecon has stronger generalization ability even with noisy viewpoint labels, deriving benefit from the manifold constraint.



\subsection{Multi-view Reconst. on OnlineProduct~\cite{song2016deep}}
\label{sec:stanfordonline}

Stanford Online Product~\cite{song2016deep} is a large-scale multiview dataset consisting of images of products from e-commerce websites. In this experiment, we test \mcrecon on multi-view real images using the network trained on the ShapeNet~\cite{shapenet} dataset with random background images from PASCAL~\cite{pascal} to make the network robust to the background noise. We visualize the results in Fig.~\ref{fig:stanfordproduct}. The result shows that our network can integrate information across multiple views of real images and reconstruct a reasonable 3D shape.

\subsection{Representation analysis}
\label{sec:anaylsis}

In this experiment, we explore the semantic expressiveness of intermediate representation of the reconstruction function $f$. Specifically, we use the intermediate representation in the recurrent neural network, which we denote as $z$, as the aggregation of multi-view observations. We use the interpolation and vector arithmetic similar to \cite{chairgen, dcgan} in the representation space of $z$.
However, unlike the above approaches, we use different modalities for the input and output which are images and 3D shapes respectively. Therefore, we can make high-level manipulation of the representation $z$ from 2D images and modify the output 3D shape. 

First, we linearly interpolate the representations from two images inter-and intra-class (Fig.~S5). We observed that the transition is smooth across various semantic properties of the 3D shapes such as length of the wing and the size of the hole on the back of the chair. Second, we extract a latent vector that contains semantic property (such as making a hole in a chair) and apply it on a different image to modify the reconstruction (Fig.~S6). Kindly refer to the supplementary material for qualitative results on these analysis.


%% file: supplementary.tex
\subsection{Implementation Details}
In this section, we cover the implementation details of our proposed network, \mcreconfull.

\begin{figure*}[h!]
    \centering
    \begin{tabular}{ ccccc } 
        \includegraphics[height=0.9\textheight]{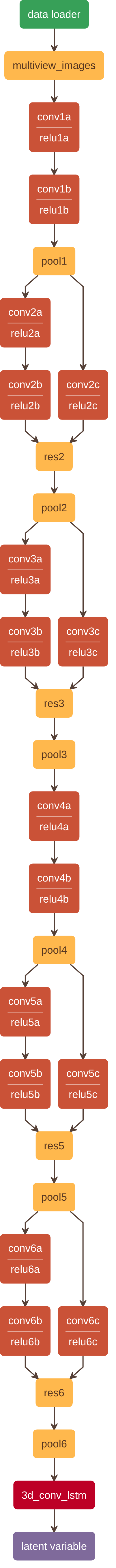}&~~~~~~~~~~&
        \includegraphics[height=0.9\textheight]{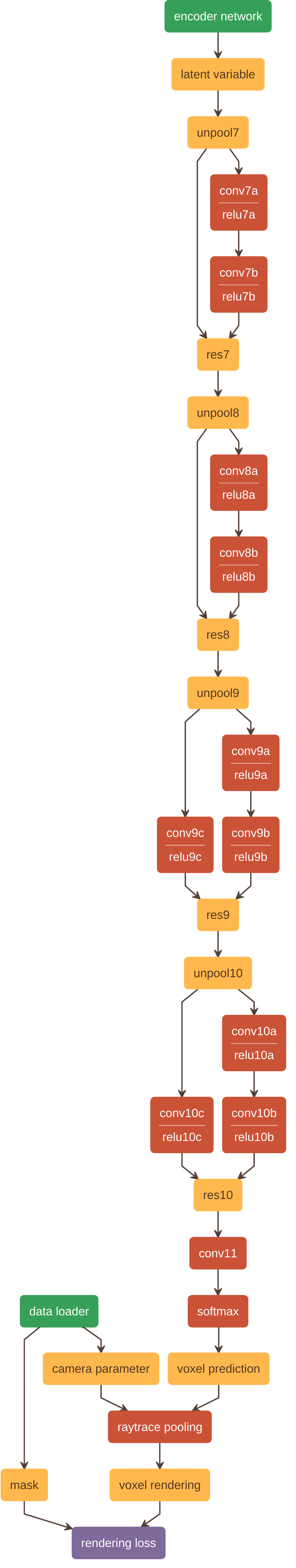}&~~~~~~~~~~&
        \includegraphics[height=0.9\textheight]{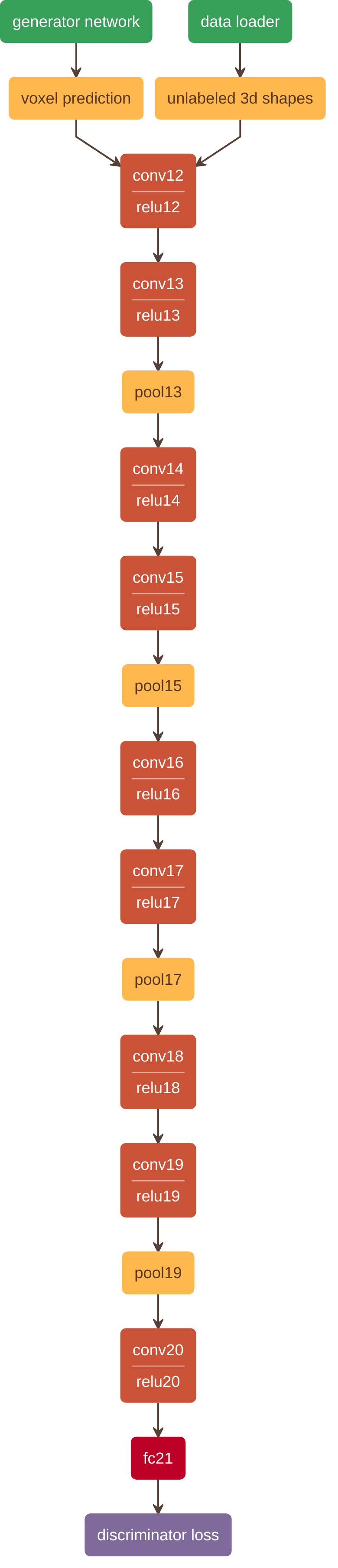}\\
        (a) encoder network && (b) generator network && (c) discriminator network
    \end{tabular}    
    \caption{Detailed network structure of \mcrecon. Please note that all of these components are connected as a single network in our implementation. We split the figure into three for better visualization.}
    \label{fig:network_supp}
\end{figure*}

\noindent \textbf{Network:} \label{supp:network}
The network is composed of three parts - an encoder, a generator, and a discriminator. The encoder and generator, following the Deep-Residual-GRU network proposed by Choy \etal\cite{3dr2n2}, learns to reconstruct volume from 2D images. The discriminator works as a  manifold constraint for weak 2D supervision. Please refer to Fig.~\ref{fig:network_supp} for detailed visualization of the network architectures. Following is the detailed description of each component of the network.

First, the encoder takes RGB image(s) $\mathbf{I}$ with size $127^2$ as an input. Each of the multi-view images is encoded into a feature vector of size 1024 through a sequence of convolutions and pooling with residual connections. The encoded feature vectors are reduced into a latent variable $z$ of size $4\times4\times4\times128$ through 3D convolutional LSTM~\cite{3dr2n2}. The first three dimensions indicate the three spatial dimensions and the last dimension indicates the feature size. The 3D convolutional LSTM works as an attention mechanism that writes features from images to corresponding voxels in 3D space. Thus, the 3D-LSTM explicitly resolves the viewpoints and self-occlusion. The encoder network is visualized in Fig. \ref{fig:network_supp} (a).

Second, as shown in Fig. \ref{fig:network_supp} (b), the generator repeats 3D convolution and unpooling until it reaches the resolution $32\times32\times32$ with residual connections like the encoder. Then, we apply one convolution followed by a softmax function to generate 3D voxel occupancy map $x$. Given the reconstruction, we compute the projection loss using the Raytrace Pooling Layer, projecting the reconstruction into a silhouette of size $127^2$.

Lastly, the discriminator takes either the reconstruction or the unlabeled shapes and generates a scalar value. The discriminator consists of a sequence of 3D convolutions and 3D max pooling until the activation is reduced to $2\times2\times2$ grid. The activation is then vectorized and fed into a fully connected layer followed by a softmax layer. Again, the network's detailed structure can be found in Fig. \ref{fig:network_supp} (c).

We implemented \mcrecon with a symbolic math neural network library~\cite{theano}. All source code and models used in this work will be publicly released upon publication.

\noindent \textbf{Optimization:} 
Training the barrier for our proposed weakly supervised reconstruction with manifold constraint faces challenges observed in prior works~\cite{dcgan, improvedtechforgan, affgan}, which we overcome using the following techniques. First, the discriminator training involves computing $\log q/p$ which can
cause divergence if support of $p$ does not overlap with the support of $q$ ($p$ is the distribution of $x^\star$ and $q$ is the distribution of $\hat{x}$).
To prevent such case, we followed the instance noise technique by S{\o}nderby \etal\cite{affgan} which smooths the probability space to make the support of $p$ infinite.
In addition, we used the update rule in \cite{3dgan} and train the
discriminator only if its prediction error becomes larger than 20\%. This technique makes the discriminator imperfect and prevents saturation of $g$. Finally,
we use different learning rate for $f$ and $g$:
$10^{-2}$ for $\theta_f$ and $10^{-4}$ for $\theta_g$ and reduce the learning
rate by the factor of 10 after 10,000 and 30,000 iterations. 
We train the network over 40,000 iterations using ADAM~\cite{adam} with batch
size 8. We used $t=100$ for all experiments.

\subsection{Baseline Methods}

\ignore{In the main paper, we conducted ablation study on our network utilizing various other baselines.}
In this section, we cover further implementation details of the baseline methods used in the main paper.

\noindent \textbf{Voxel Carving (VC):} Given silhouettes and camera parameters, voxel carving~\cite{spacecarving} removes voxels that lie outside of the silhouettes when projected to the image planes. Please note that voxel carving always requires camera parameters and masks, in contrast to all other learning-based methods which only require an image as an input.

\noindent \textbf{Raytrace Pooling (RP):} We train an encoder-generator network only with mask
supervisions ($\mathcal{L}_{reproj}$). The network has the same architecture as the \mcrecon as shown in \ref{supp:network} but does not have a discriminator that provides gradients toward 3D shape manifold. Please note that the mask supervision requires Raytrace Pooling that we proposed.

\noindent \textbf{Perspective Transformer (PT):} \cite{singleviewproj} proposed a perspective projection layer (Perspective Transformer) that is similar to the RP Layer. To compare it with the RP Layer, we propose another baseline, an encoder-generator network only with mask supervisions, but with the Perspective Transformer (PT). Since the base network architecture affects the performance drastically, we use the same network for all learning based methods including this one. While the RP uses an accurate raytracing, the PT uses sampling points from a 3D grid over a fixed range of depth from camera center on the voxel space. Therefore, the PT requires hyperparameters for the range and the density of the samples. We determined the range by experimentally measuring the minimal and maximal possible depth of the voxel space over the training data and used sampling density 16 by default as suggested by \cite{singleviewproj}. Additionally, we vary the density of the sample to measure the effect of the sampling at main paper Sec. 4.3.
 
\noindent \textbf{Raytrace Pooling + Shape Nearest Neighbor (RP + NN):} For a simple baseline that uses both unlabeled 3D shapes and 2D weak supervision, we propose a nearest neighbor retrieval of the unlabeled 3D shapes with RP. We first use the RP network to generate prediction and retrieve the nearest neighbor within the unlabeled 3D shapes. This method improves prediction accuracy if there is a similar shape among the unlabeled 3D shapes and the prediction from the RP network is accurate.

\ignore{As an alternative to interpolate over unlabeled 3D shapes $\hat{x}$, we propose another baseline experiment - nearest neighbor retrieval(NN). For this experiment, we retrieve a nearest neighbor 3D shape of the output of RP among the unlabeled 3D shapes. Please note that approach requires unlabeled 3D shapes $\hat{x}$ at test time, unlike \mcrecon.}

\noindent \textbf{Full 3D Supervision (F3D):} Finally, we provide the results from full 3D supervision~\cite{3dr2n2} as reference. The networks are trained with 3D supervision (3D shapes) on the same network architecture as in \ref{supp:network} without discriminator providing manifold constraint. This experiment provides an upper bound performance for our \mcrecon since 2D projections only provide partial information of the 3D shapes.

\subsection{Single-view reconst. on IKEA dataset\cite{ikea}}

In order to compare our work with the other recent supervised 3D reconstruction methods~\cite{tlnet,3dgan}, we tested our network on IKEA dataset. Similar to other works, we trained a single network on ShapeNet renderings of the furniture merged with random background from PASCAL~\cite{pascal}. Following the convention of \cite{tlnet,3dgan}, we evaluated the reconstruction on ground-truth model aligned over permutations, flips, and translational alignments (up to 10\%). Please note that all of the other baselines require full 3D supervision that is meant to provide upper bound performance over \mcrecon. The quantitative results can be found in Tab. \ref{tab:ikea}.

\begin{table}[ht]
    \scriptsize
    \centering
    \begin{tabular}{| c | c c c c | c |}
    \hline
    Method & Chair & Desk & Sofa & Table & Mean\\ \hline
    AlexNet-fc8\cite{tlnet} & 20.4 & 19.7 & 38.8 & 16.0 & 23.7\\
    AlexNet-conf4\cite{tlnet} & 31.4 & 26.6 & 69.3 & 19.1 & 37.1\\
    T-L Network\cite{tlnet} & 32.9 & 25.8 & 71.7 & 23.3 & 39.6\\
    3D-VAE-GAN\cite{3dgan} & 42.6 & 34.8 & 79.8 & 33.1 & 48.8\\ \hline
    \mcrecon & 32.0 & 28.6 & 55.7 & 29.0 & 37.0 \\\hline
    \end{tabular}
    \vspace{0.3cm}
    \caption{Per-class real image 3D reconstruction Average Precision(AP) percentage on IKEA dataset\cite{ikea}. Please note that all of the other baselines require full 3D supervision that are meant to provide upper bound performance over \mcrecon.}
    \label{tab:ikea}
\end{table}




\subsection{Multi-view synthetic images reconstruction}
\begin{figure*}[h!]
    \small
    \centering
    \begin{tabular}{ccccccccc}
    \toprule
    Input &
    \includegraphics[height=0.08\linewidth,width=0.08\linewidth,keepaspectratio]{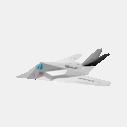} &
    \includegraphics[height=0.08\linewidth,width=0.08\linewidth,keepaspectratio]{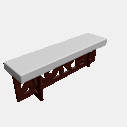} &
    \includegraphics[height=0.08\linewidth,width=0.08\linewidth,keepaspectratio]{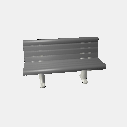} &
    \includegraphics[height=0.08\linewidth,width=0.08\linewidth,keepaspectratio]{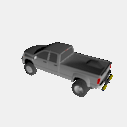} &
    \includegraphics[height=0.08\linewidth,width=0.08\linewidth,keepaspectratio]{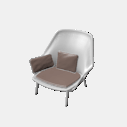} &
    \includegraphics[height=0.08\linewidth,width=0.08\linewidth,keepaspectratio]{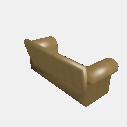} &
    \includegraphics[height=0.08\linewidth,width=0.08\linewidth,keepaspectratio]{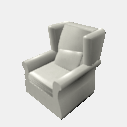} &
    \includegraphics[height=0.08\linewidth,width=0.08\linewidth,keepaspectratio]{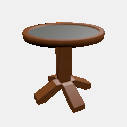} \\
    \midrule
    \specialcell{G.T.}&
    \includegraphics[height=0.08\linewidth,width=0.08\linewidth,keepaspectratio]{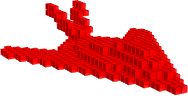} &
    \includegraphics[height=0.08\linewidth,width=0.08\linewidth,keepaspectratio]{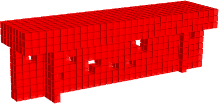} &
    \includegraphics[height=0.08\linewidth,width=0.08\linewidth,keepaspectratio]{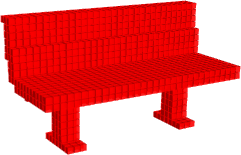} &
    \includegraphics[height=0.08\linewidth,width=0.08\linewidth,keepaspectratio]{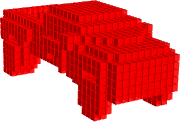} &
    \includegraphics[height=0.08\linewidth,width=0.08\linewidth,keepaspectratio]{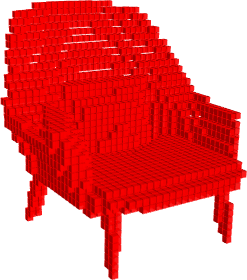} &
    \includegraphics[height=0.08\linewidth,width=0.08\linewidth,keepaspectratio]{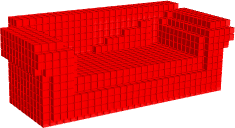} &
    \includegraphics[height=0.08\linewidth,width=0.08\linewidth,keepaspectratio]{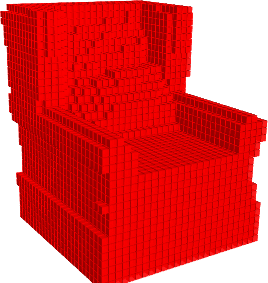} &
    \includegraphics[height=0.08\linewidth,width=0.08\linewidth,keepaspectratio]{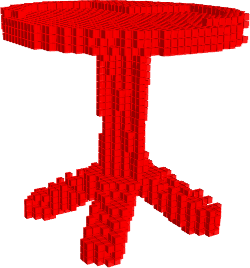} \\
    \midrule
    \specialcell{1 view\\RP}&
    \includegraphics[height=0.08\linewidth,width=0.08\linewidth,keepaspectratio]{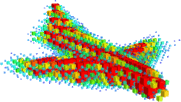} &
    \includegraphics[height=0.08\linewidth,width=0.08\linewidth,keepaspectratio]{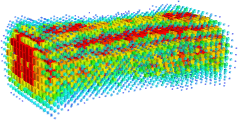} &
    \includegraphics[height=0.08\linewidth,width=0.08\linewidth,keepaspectratio]{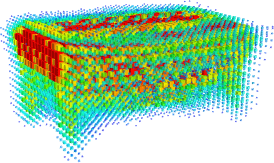} &
    \includegraphics[height=0.08\linewidth,width=0.08\linewidth,keepaspectratio]{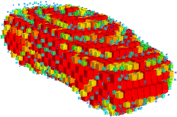} &
    \includegraphics[height=0.08\linewidth,width=0.08\linewidth,keepaspectratio]{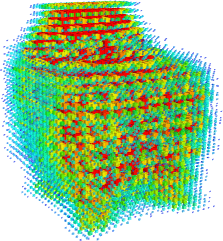} &
    \includegraphics[height=0.08\linewidth,width=0.08\linewidth,keepaspectratio]{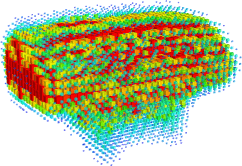} &
    \includegraphics[height=0.08\linewidth,width=0.08\linewidth,keepaspectratio]{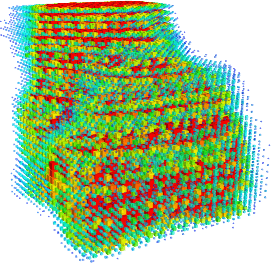} &
    \includegraphics[height=0.08\linewidth,width=0.08\linewidth,keepaspectratio]{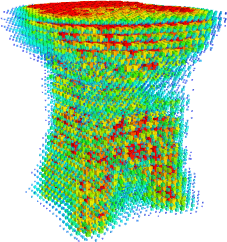} \\
    \specialcell{1 view\\ \mcrecon} &
    \includegraphics[height=0.08\linewidth,width=0.08\linewidth,keepaspectratio]{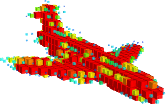} &
    \includegraphics[height=0.08\linewidth,width=0.08\linewidth,keepaspectratio]{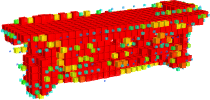} &
    \includegraphics[height=0.08\linewidth,width=0.08\linewidth,keepaspectratio]{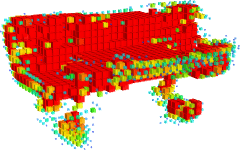} &
    \includegraphics[height=0.08\linewidth,width=0.08\linewidth,keepaspectratio]{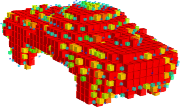} &
    \includegraphics[height=0.08\linewidth,width=0.08\linewidth,keepaspectratio]{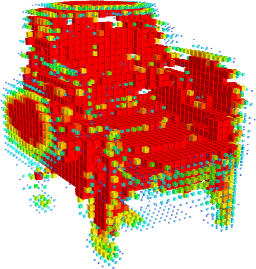} &
    \includegraphics[height=0.08\linewidth,width=0.08\linewidth,keepaspectratio]{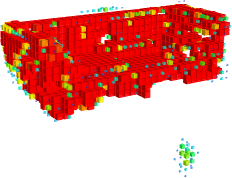} &
    \includegraphics[height=0.08\linewidth,width=0.08\linewidth,keepaspectratio]{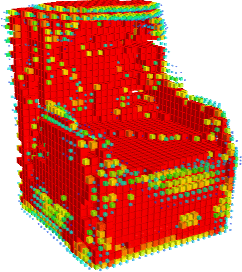} &
    \includegraphics[height=0.08\linewidth,width=0.08\linewidth,keepaspectratio]{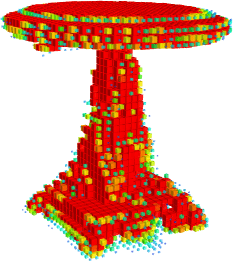} \\
    \midrule
    \specialcell{5 views\\ RP}&
    \includegraphics[height=0.08\linewidth,width=0.08\linewidth,keepaspectratio]{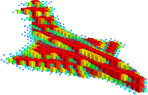} &
    \includegraphics[height=0.08\linewidth,width=0.08\linewidth,keepaspectratio]{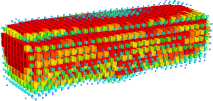} &
    \includegraphics[height=0.08\linewidth,width=0.08\linewidth,keepaspectratio]{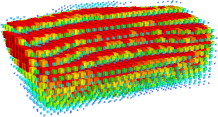} &
    \includegraphics[height=0.08\linewidth,width=0.08\linewidth,keepaspectratio]{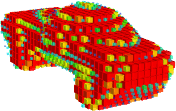} &
    \includegraphics[height=0.08\linewidth,width=0.08\linewidth,keepaspectratio]{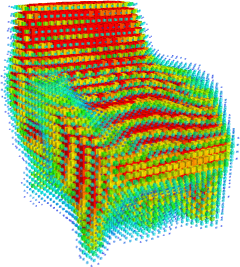} &
    \includegraphics[height=0.08\linewidth,width=0.08\linewidth,keepaspectratio]{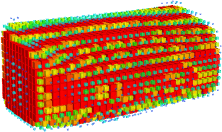} &
    \includegraphics[height=0.08\linewidth,width=0.08\linewidth,keepaspectratio]{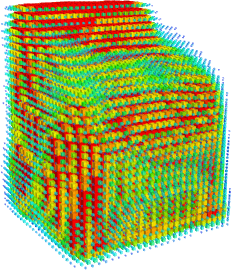} &
    \includegraphics[height=0.08\linewidth,width=0.08\linewidth,keepaspectratio]{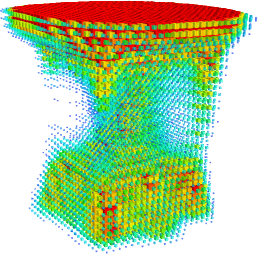} \\
    \specialcell{5 views\\ \mcrecon}&
    \includegraphics[height=0.08\linewidth,width=0.08\linewidth,keepaspectratio]{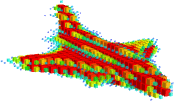} &
    \includegraphics[height=0.08\linewidth,width=0.08\linewidth,keepaspectratio]{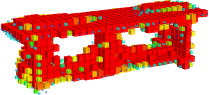} &
    \includegraphics[height=0.08\linewidth,width=0.08\linewidth,keepaspectratio]{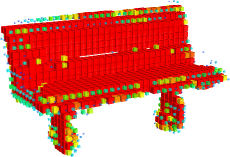} &
    \includegraphics[height=0.08\linewidth,width=0.08\linewidth,keepaspectratio]{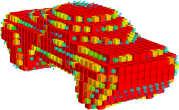} &
    \includegraphics[height=0.08\linewidth,width=0.08\linewidth,keepaspectratio]{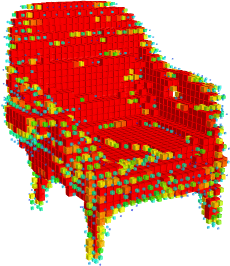} &
    \includegraphics[height=0.08\linewidth,width=0.08\linewidth,keepaspectratio]{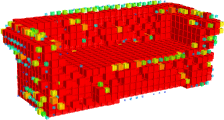} &
    \includegraphics[height=0.08\linewidth,width=0.08\linewidth,keepaspectratio]{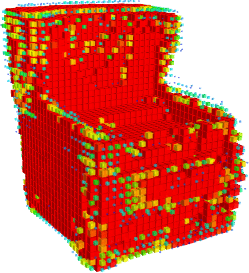} &
    \includegraphics[height=0.08\linewidth,width=0.08\linewidth,keepaspectratio]{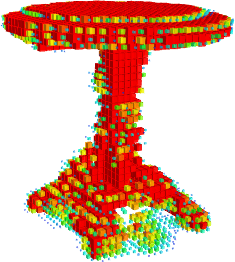} \\
    \bottomrule
    \multicolumn{9}{c}{(a)}\\
    \toprule
    Input &
    \includegraphics[height=0.08\linewidth,width=0.08\linewidth,keepaspectratio]{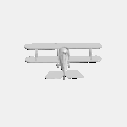} &
    \includegraphics[height=0.08\linewidth,width=0.08\linewidth,keepaspectratio]{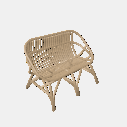} &
    \includegraphics[height=0.08\linewidth,width=0.08\linewidth,keepaspectratio]{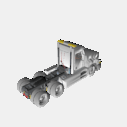} &
    \includegraphics[height=0.08\linewidth,width=0.08\linewidth,keepaspectratio]{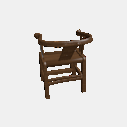} &
    \includegraphics[height=0.08\linewidth,width=0.08\linewidth,keepaspectratio]{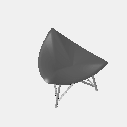} &
    \includegraphics[height=0.08\linewidth,width=0.08\linewidth,keepaspectratio]{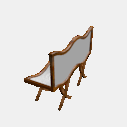} &
    \includegraphics[height=0.08\linewidth,width=0.08\linewidth,keepaspectratio]{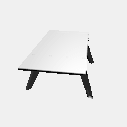} &
    \includegraphics[height=0.08\linewidth,width=0.08\linewidth,keepaspectratio]{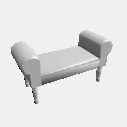} \\
    \midrule
    \specialcell{G.T.}&
    \includegraphics[height=0.08\linewidth,width=0.08\linewidth,keepaspectratio]{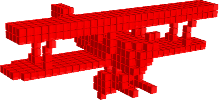} &
    \includegraphics[height=0.08\linewidth,width=0.08\linewidth,keepaspectratio]{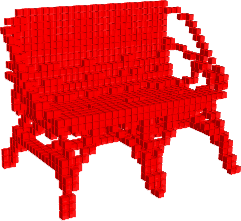} &
    \includegraphics[height=0.08\linewidth,width=0.08\linewidth,keepaspectratio]{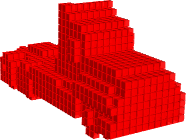} &
    \includegraphics[height=0.08\linewidth,width=0.08\linewidth,keepaspectratio]{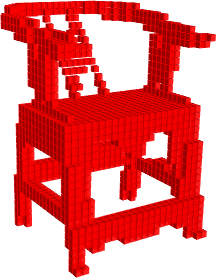} &
    \includegraphics[height=0.08\linewidth,width=0.08\linewidth,keepaspectratio]{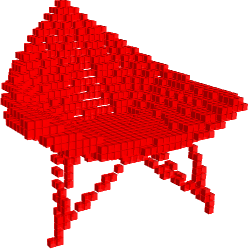} &
    \includegraphics[height=0.08\linewidth,width=0.08\linewidth,keepaspectratio]{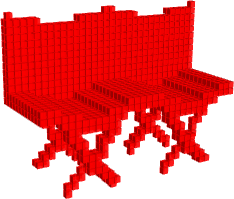} &
    \includegraphics[height=0.08\linewidth,width=0.08\linewidth,keepaspectratio]{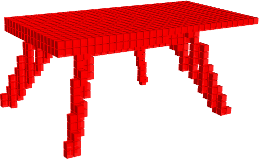} &
    \includegraphics[height=0.08\linewidth,width=0.08\linewidth,keepaspectratio]{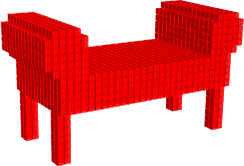} \\
    \midrule
    \specialcell{1 view\\RP}&
    \includegraphics[height=0.08\linewidth,width=0.08\linewidth,keepaspectratio]{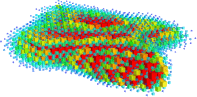} &
    \includegraphics[height=0.08\linewidth,width=0.08\linewidth,keepaspectratio]{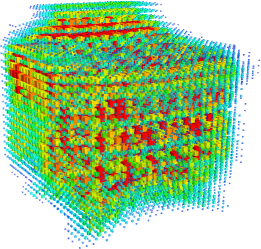} &
    \includegraphics[height=0.08\linewidth,width=0.08\linewidth,keepaspectratio]{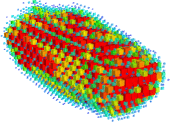} &
    \includegraphics[height=0.08\linewidth,width=0.08\linewidth,keepaspectratio]{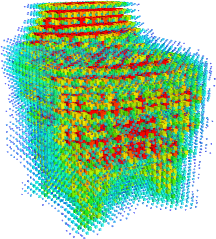} &
    \includegraphics[height=0.08\linewidth,width=0.08\linewidth,keepaspectratio]{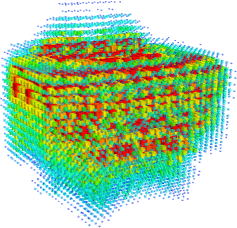} &
    \includegraphics[height=0.08\linewidth,width=0.08\linewidth,keepaspectratio]{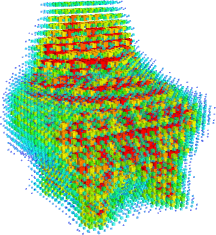} &
    \includegraphics[height=0.08\linewidth,width=0.08\linewidth,keepaspectratio]{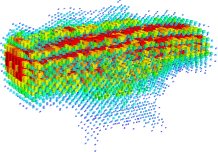} &
    \includegraphics[height=0.08\linewidth,width=0.08\linewidth,keepaspectratio]{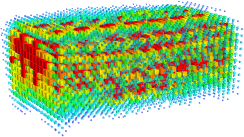} \\
    \specialcell{1 view\\ \mcrecon} &
    \includegraphics[height=0.08\linewidth,width=0.08\linewidth,keepaspectratio]{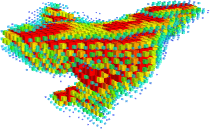} &
    \includegraphics[height=0.08\linewidth,width=0.08\linewidth,keepaspectratio]{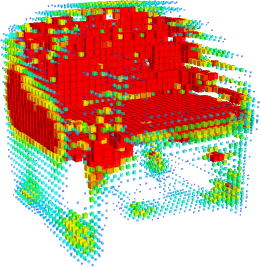} &
    \includegraphics[height=0.08\linewidth,width=0.08\linewidth,keepaspectratio]{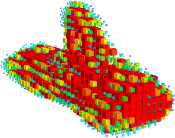} &
    \includegraphics[height=0.08\linewidth,width=0.08\linewidth,keepaspectratio]{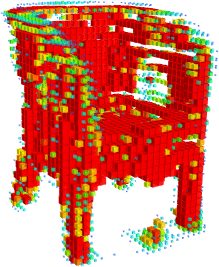} &
    \includegraphics[height=0.08\linewidth,width=0.08\linewidth,keepaspectratio]{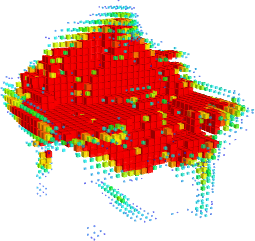} &
    \includegraphics[height=0.08\linewidth,width=0.08\linewidth,keepaspectratio]{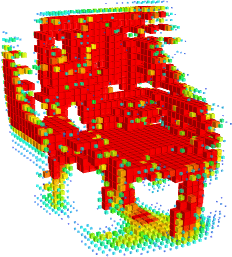} &
    \includegraphics[height=0.08\linewidth,width=0.08\linewidth,keepaspectratio]{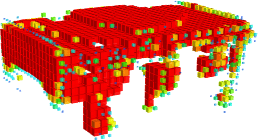} &
    \includegraphics[height=0.08\linewidth,width=0.08\linewidth,keepaspectratio]{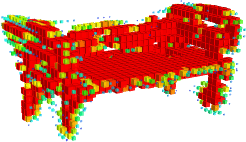} \\
    \midrule
    \specialcell{5 views\\RP}&
    \includegraphics[height=0.08\linewidth,width=0.08\linewidth,keepaspectratio]{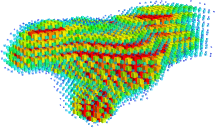} &
    \includegraphics[height=0.08\linewidth,width=0.08\linewidth,keepaspectratio]{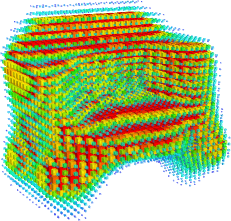} &
    \includegraphics[height=0.08\linewidth,width=0.08\linewidth,keepaspectratio]{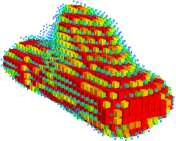} &
    \includegraphics[height=0.08\linewidth,width=0.08\linewidth,keepaspectratio]{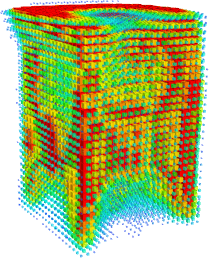} &
    \includegraphics[height=0.08\linewidth,width=0.08\linewidth,keepaspectratio]{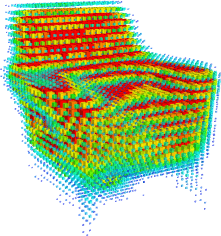} &
    \includegraphics[height=0.08\linewidth,width=0.08\linewidth,keepaspectratio]{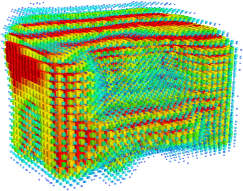} &
    \includegraphics[height=0.08\linewidth,width=0.08\linewidth,keepaspectratio]{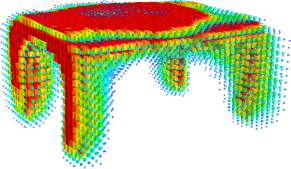} &
    \includegraphics[height=0.08\linewidth,width=0.08\linewidth,keepaspectratio]{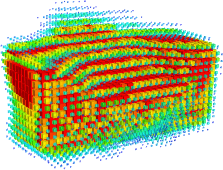} \\
    \specialcell{5 views\\ \mcrecon}&
    \includegraphics[height=0.08\linewidth,width=0.08\linewidth,keepaspectratio]{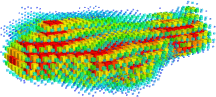} &
    \includegraphics[height=0.08\linewidth,width=0.08\linewidth,keepaspectratio]{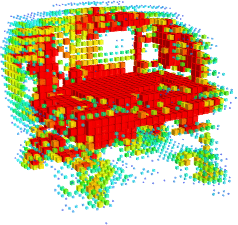} &
    \includegraphics[height=0.08\linewidth,width=0.08\linewidth,keepaspectratio]{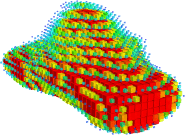} &
    \includegraphics[height=0.08\linewidth,width=0.08\linewidth,keepaspectratio]{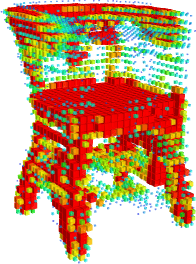} &
    \includegraphics[height=0.08\linewidth,width=0.08\linewidth,keepaspectratio]{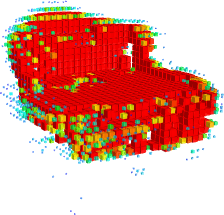} &
    \includegraphics[height=0.08\linewidth,width=0.08\linewidth,keepaspectratio]{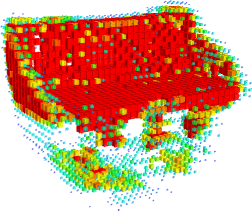} &
    \includegraphics[height=0.08\linewidth,width=0.08\linewidth,keepaspectratio]{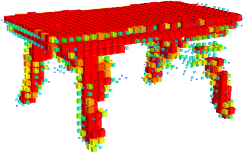} &
    \includegraphics[height=0.08\linewidth,width=0.08\linewidth,keepaspectratio]{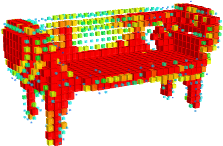} \\
    \bottomrule
    \multicolumn{9}{c}{(b)}
    \end{tabular}
\vspace{0.3cm}
\caption{(a) Successful (b) less-successful qualitative results of single- or multi-view synthetic image reconstructions on ShapeNet dataset. This result hints that our \mcrecon~is learning high-quality reconstruction including concavity from a small number of views of mask supervision. Please check the main paper for details of our visualization method.}
    \label{fig:shapenet-supp}
\end{figure*}

In Figure \ref{fig:shapenet-supp}, we visualized more qualitative reconstruction results on ShapeNet~\cite{shapenet} dataset. In order to visualize the strength and the weakness of \mcrecon, we presented both successful and less-successful reconstruction results. In general, as discussed in the main paper, \mcrecon reconstructed a reasonable 3D shape from a small number of silhouettes and viewpoints. However, \mcrecon had some difficulty reconstructing exotic shapes which might not be in the unlabeled shape repository given to the discriminator to be learned as a target shape manifold.

\subsection{Single real image reconstruction}
\begin{figure*}[ht!]
    \small
    \centering
    \setlength\extrarowheight{2pt}
    \begin{tabular}{ccccccccc}
    \toprule
    Input &
    \includegraphics[height=0.08\linewidth,width=0.08\linewidth,keepaspectratio]{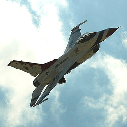} &
    \includegraphics[height=0.08\linewidth,width=0.08\linewidth,keepaspectratio]{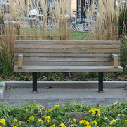} &
    \includegraphics[height=0.08\linewidth,width=0.08\linewidth,keepaspectratio]{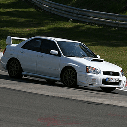} &
    \includegraphics[height=0.08\linewidth,width=0.08\linewidth,keepaspectratio]{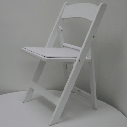} &
    \includegraphics[height=0.08\linewidth,width=0.08\linewidth,keepaspectratio]{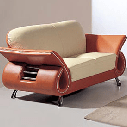} &
    \includegraphics[height=0.08\linewidth,width=0.08\linewidth,keepaspectratio]{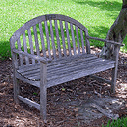} &
    \includegraphics[height=0.08\linewidth,width=0.08\linewidth,keepaspectratio]{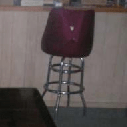} &
    \includegraphics[height=0.08\linewidth,width=0.08\linewidth,keepaspectratio]{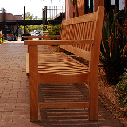} \\
    \midrule
    \specialcell{G.T.}&
    \includegraphics[height=0.08\linewidth,width=0.08\linewidth,keepaspectratio]{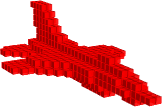} &
    \includegraphics[height=0.08\linewidth,width=0.08\linewidth,keepaspectratio]{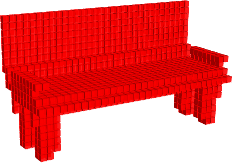} &
    \includegraphics[height=0.08\linewidth,width=0.08\linewidth,keepaspectratio]{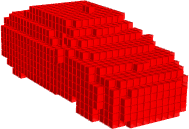} &
    \includegraphics[height=0.08\linewidth,width=0.08\linewidth,keepaspectratio]{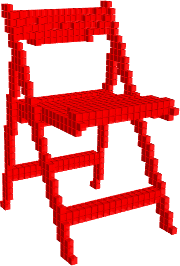} &
    \includegraphics[height=0.08\linewidth,width=0.08\linewidth,keepaspectratio]{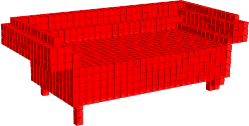} &
    \includegraphics[height=0.08\linewidth,width=0.08\linewidth,keepaspectratio]{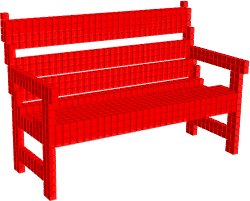} &
    \includegraphics[height=0.08\linewidth,width=0.08\linewidth,keepaspectratio]{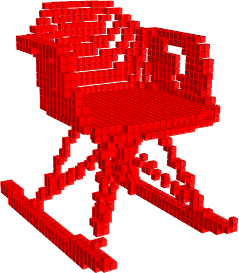} &
    \includegraphics[height=0.08\linewidth,width=0.08\linewidth,keepaspectratio]{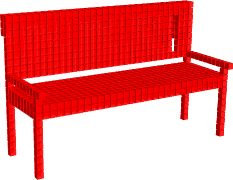} \\
    \midrule
    \specialcell{VC}&
    \includegraphics[height=0.08\linewidth,width=0.08\linewidth,keepaspectratio]{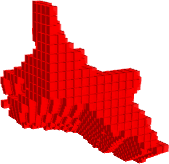} &
    \includegraphics[height=0.08\linewidth,width=0.08\linewidth,keepaspectratio]{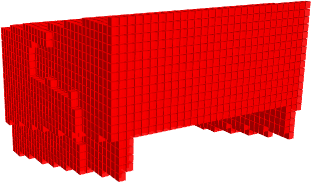} &
    \includegraphics[height=0.08\linewidth,width=0.08\linewidth,keepaspectratio]{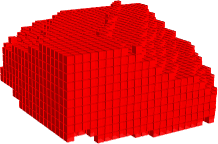} &
    \includegraphics[height=0.08\linewidth,width=0.08\linewidth,keepaspectratio]{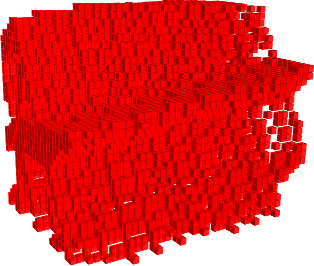} &
    \includegraphics[height=0.08\linewidth,width=0.08\linewidth,keepaspectratio]{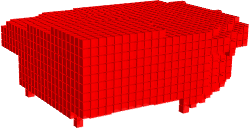} &
    \includegraphics[height=0.08\linewidth,width=0.08\linewidth,keepaspectratio]{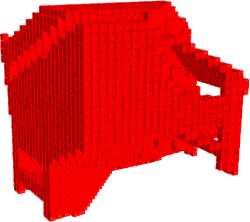} &
    \includegraphics[height=0.08\linewidth,width=0.08\linewidth,keepaspectratio]{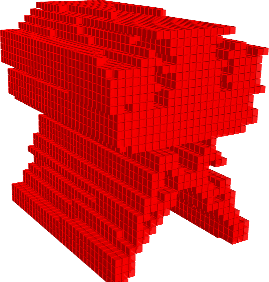} &
    \includegraphics[height=0.08\linewidth,width=0.08\linewidth,keepaspectratio]{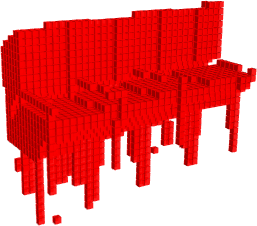} \\
    \mcrecon &
    \includegraphics[height=0.08\linewidth,width=0.08\linewidth,keepaspectratio]{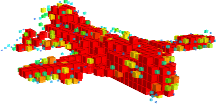} &
    \includegraphics[height=0.08\linewidth,width=0.08\linewidth,keepaspectratio]{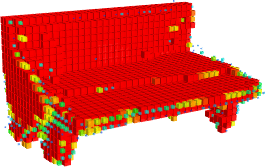} &
    \includegraphics[height=0.08\linewidth,width=0.08\linewidth,keepaspectratio]{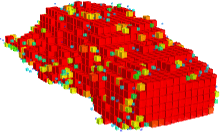} &
    \includegraphics[height=0.08\linewidth,width=0.08\linewidth,keepaspectratio]{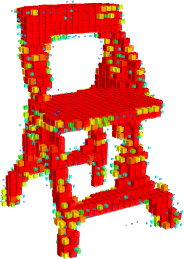} &
    \includegraphics[height=0.08\linewidth,width=0.08\linewidth,keepaspectratio]{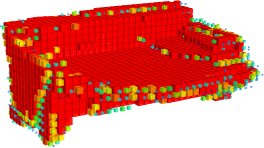} &
    \includegraphics[height=0.08\linewidth,width=0.08\linewidth,keepaspectratio]{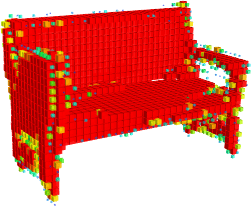} &
    \includegraphics[height=0.08\linewidth,width=0.08\linewidth,keepaspectratio]{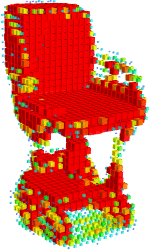} &
    \includegraphics[height=0.08\linewidth,width=0.08\linewidth,keepaspectratio]{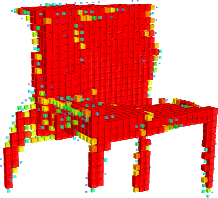} \\
    \bottomrule
    \end{tabular}
\vspace{0.3cm}
    \caption{Qualitative results of real image reconstructions on ObjectNet3D. The results hints that our network successfully carved out concavity, which cannot be learned from mask supervision. Please note that voxel carving requires camera parameter at test time while ours does not.}
    \label{fig:objectnet3d-supp}
\end{figure*}

In Figure~\ref{fig:objectnet3d-supp}, we visualized more qualitative reconstruction results on ObjectNet3D\cite{objectnet3d} dataset. We observed that \mcrecon can learn to reconstruct a reasonable 3D shape from a single mask supervision.

\subsection{Multi-view real image reconstruction}
\begin{figure*}
    \centering
    \setlength\extrarowheight{2pt}
    \begin{tabular}{cccccccccc}
    \toprule
    Input &
    \includegraphics[height=0.08\linewidth,width=0.08\linewidth,keepaspectratio]{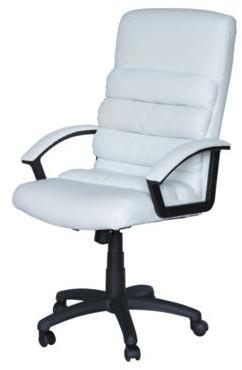} &
    \includegraphics[height=0.08\linewidth,width=0.08\linewidth,keepaspectratio]{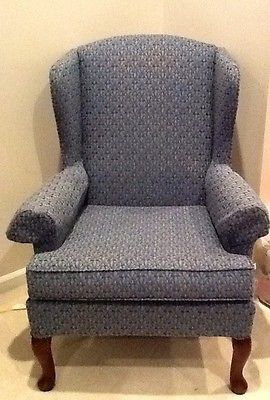} &
    \includegraphics[height=0.08\linewidth,width=0.08\linewidth,keepaspectratio]{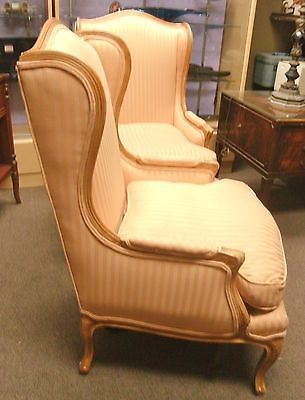} &
    \includegraphics[height=0.08\linewidth,width=0.08\linewidth,keepaspectratio]{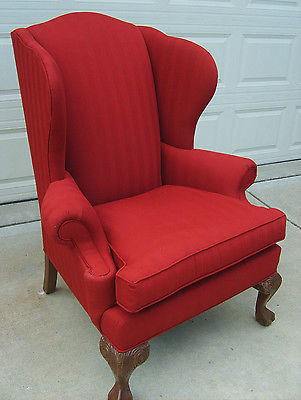} &
    \includegraphics[height=0.08\linewidth,width=0.08\linewidth,keepaspectratio]{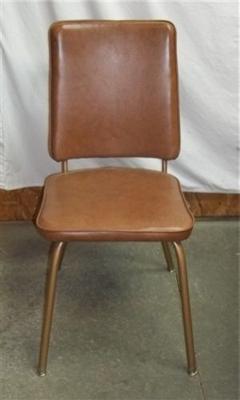} &
    \includegraphics[height=0.08\linewidth,width=0.08\linewidth,keepaspectratio]{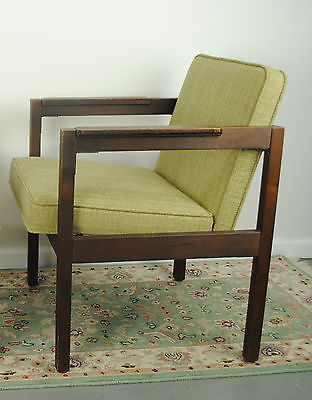} &
    \includegraphics[height=0.08\linewidth,width=0.08\linewidth,keepaspectratio]{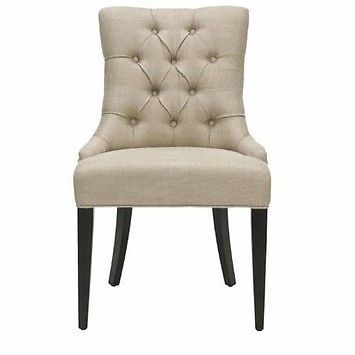} &
    \includegraphics[height=0.08\linewidth,width=0.08\linewidth,keepaspectratio]{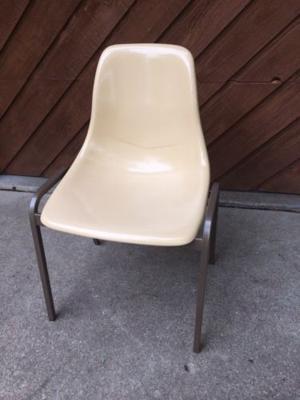} &
    \includegraphics[height=0.08\linewidth,width=0.08\linewidth,keepaspectratio]{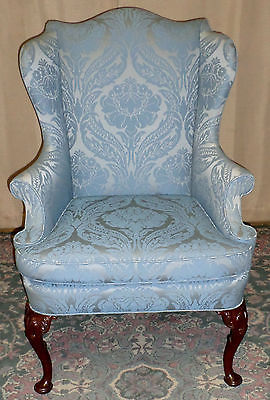} \\
    &
    \includegraphics[height=0.08\linewidth,width=0.08\linewidth,keepaspectratio]{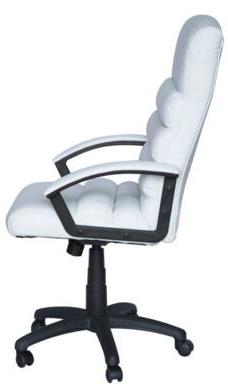} &
    \includegraphics[height=0.08\linewidth,width=0.08\linewidth,keepaspectratio]{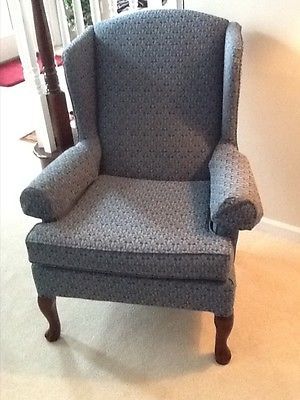} &
    \includegraphics[height=0.08\linewidth,width=0.08\linewidth,keepaspectratio]{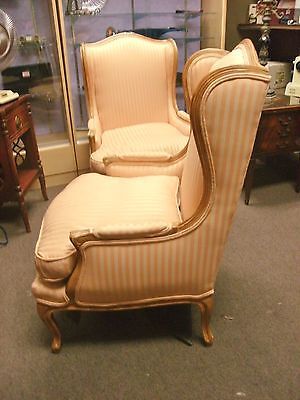} &
    \includegraphics[height=0.08\linewidth,width=0.08\linewidth,keepaspectratio]{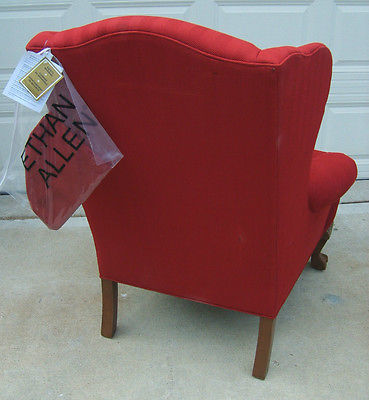} &
    \includegraphics[height=0.08\linewidth,width=0.08\linewidth,keepaspectratio]{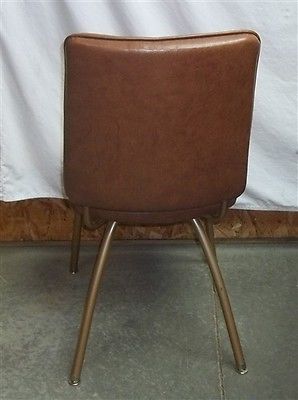} &
    \includegraphics[height=0.08\linewidth,width=0.08\linewidth,keepaspectratio]{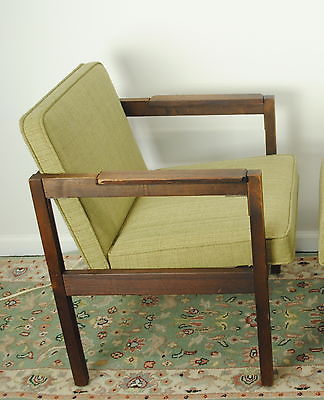} &
    \includegraphics[height=0.08\linewidth,width=0.08\linewidth,keepaspectratio]{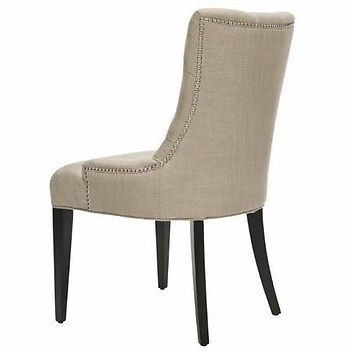} &
    \includegraphics[height=0.08\linewidth,width=0.08\linewidth,keepaspectratio]{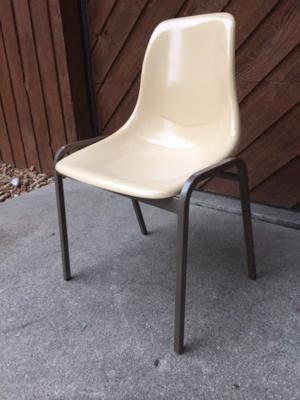} &
    \includegraphics[height=0.08\linewidth,width=0.08\linewidth,keepaspectratio]{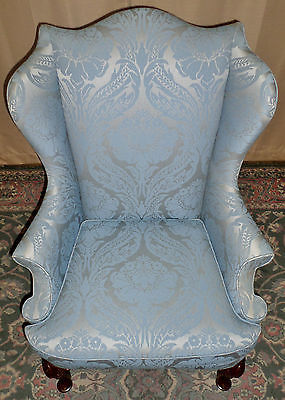} \\
    \midrule
    \mcrecon &
    \includegraphics[height=0.08\linewidth,width=0.08\linewidth,keepaspectratio]{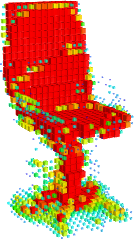} &
    \includegraphics[height=0.08\linewidth,width=0.08\linewidth,keepaspectratio]{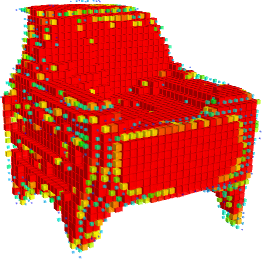} &
    \includegraphics[height=0.08\linewidth,width=0.08\linewidth,keepaspectratio]{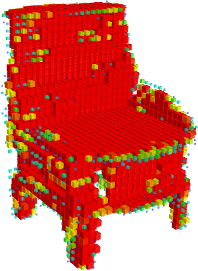} &
    \includegraphics[height=0.08\linewidth,width=0.08\linewidth,keepaspectratio]{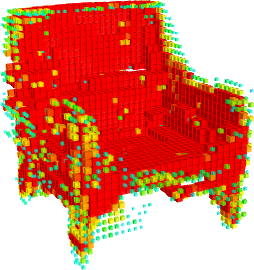} &
    \includegraphics[height=0.08\linewidth,width=0.08\linewidth,keepaspectratio]{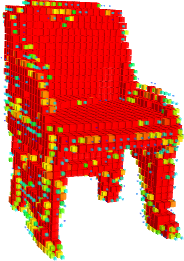} &
    \includegraphics[height=0.08\linewidth,width=0.08\linewidth,keepaspectratio]{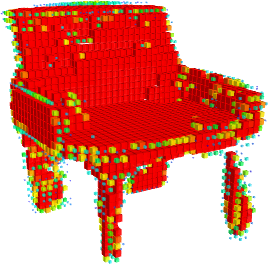} &
    \includegraphics[height=0.08\linewidth,width=0.08\linewidth,keepaspectratio]{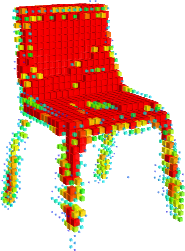} &
    \includegraphics[height=0.08\linewidth,width=0.08\linewidth,keepaspectratio]{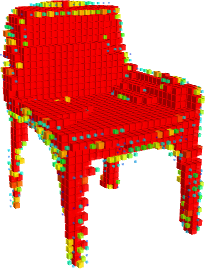} &
    \includegraphics[height=0.08\linewidth,width=0.08\linewidth,keepaspectratio]{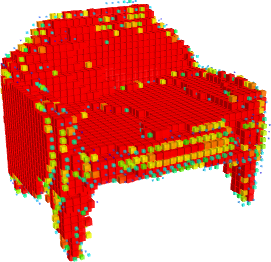} \\
    \bottomrule
    \end{tabular}
    \caption{Qualitative results of multi-view real image reconstructions on Stanford Online Product dataset~\cite{song2016deep}. Our network successfully reconstructed real images coordinating multi-view information. Please note that the domain of training is different from that of test, which makes the reconstruction more challenging.}
    \label{fig:stanfordproduct-supp}
\end{figure*}

In Figure~\ref{fig:stanfordproduct-supp}, we visualized more qualitative reconstruction results on Stanford Online Product Dataset~\cite{song2016deep}. As explained in the main paper, we trained the network on the ShapeNet~\cite{shapenet} dataset with random background images from PASCAL~\cite{pascal} to make the network robust to the background noise. Since the domain of the train and test data are different, the reconstruction quality may not be as good as other experiments. However, our network shows reasonable 3D reconstruction results.

\subsection{Representation analysis}
\begin{figure*}
    \centering
    \begin{tabular}{cc}
    \includegraphics[width=0.45\linewidth,keepaspectratio]{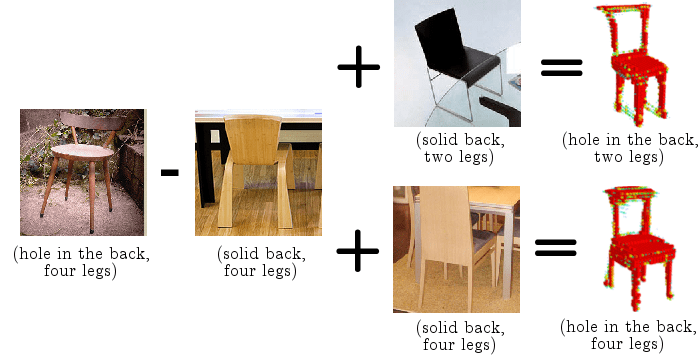} &
    \includegraphics[width=0.45\linewidth,keepaspectratio]{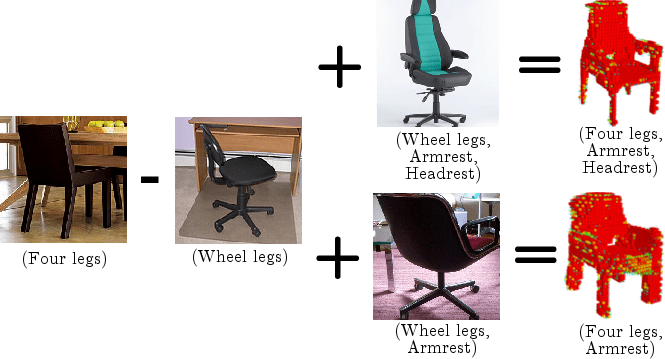} \\
    \includegraphics[width=0.45\linewidth,keepaspectratio]{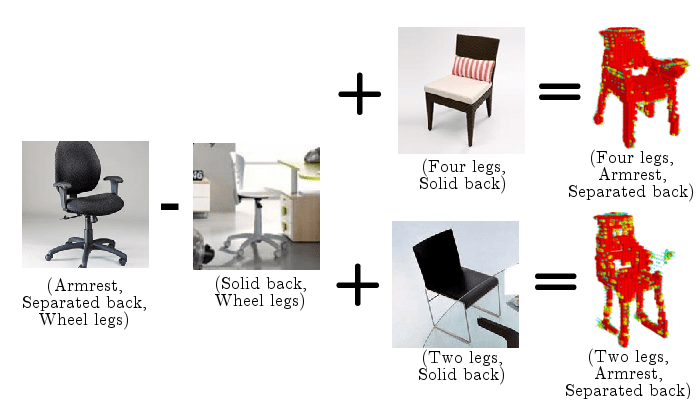} &
    \includegraphics[width=0.45\linewidth,keepaspectratio]{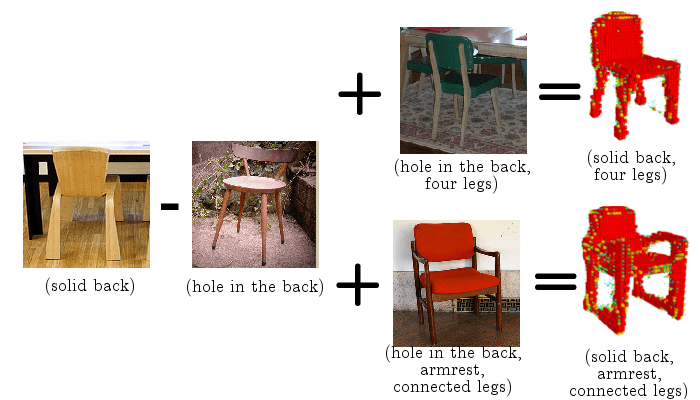} \\
    \includegraphics[width=0.45\linewidth,keepaspectratio]{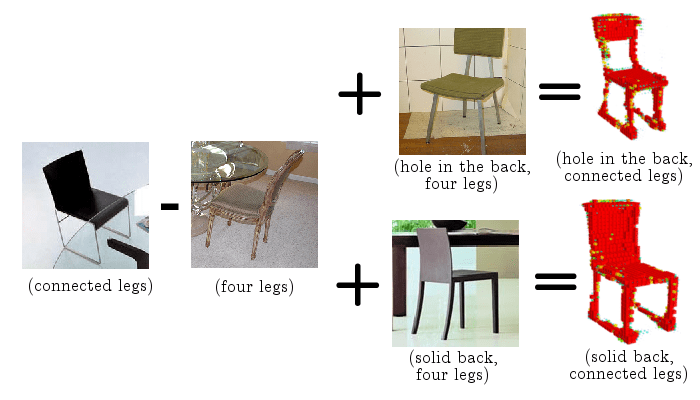} &
    \includegraphics[width=0.45\linewidth,keepaspectratio]{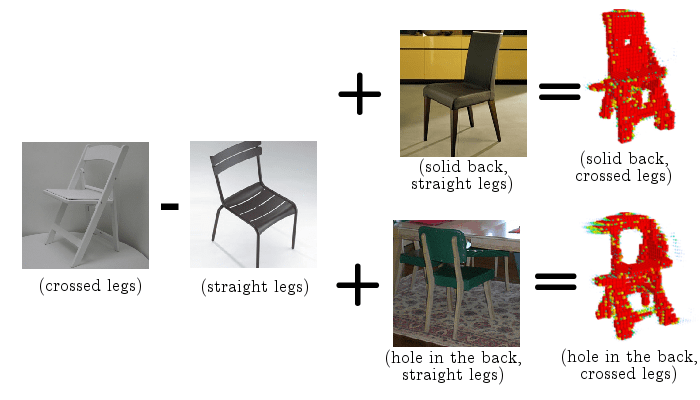} \\
    \end{tabular}
    \caption{Arithmetic on latent variable $z$ of different images. By subtracting latent variables of similar chairs with different properties, we extracted the feature which represents such property. We applied the feature to two other chairs to demonstrate that this is a generic and replicable representation}
    \label{fig:arithmetic-supp}
\end{figure*}

We present more representation analysis results similar to the results in the main paper Sec. 4.6. \ignore{Following the experiments from the main paper, we visualize further results of our representation analysis.}
In Figure~\ref{fig:interpolation-supp}, we linearly interpolate the latent variables of two images inter-and intra-class. This shows that the latent space that the encoder learned is the smooth space over the 3D shapes. In Figure~\ref{fig:arithmetic-supp}, we add and subtract the latent variables of different images to modify the generated voxels with semantic context. Both experiments hint that the latent variable of \mcrecon has a meaningful semantic expressiveness that allows us to manipulate 3D shapes semantically.

\subsection{Computation Time}
We evaluated computation time of all methods in our experiments. All experiments are on NVIDIA Titan X with batch size 8 and 5 views. Please note that at test time, we do not need to evaluate the manifold projecting discriminator, thus, the computation time is the same for RP and \mcrecon.
\begin{center}
\centering
\scriptsize
\begin{tabular}{ |c|c|c|c|c| } 
 \hline
 Method & Voxel Carving  & RP train &  \mcrecon train &  \mcrecon test\\ \hline
 Time(s) & 0.115  & 3.57 & 5.16 & 0.268 \\ \hline
\end{tabular}
\end{center}

\newcommand{\hb}[0]{0.08}
\newcommand{\wb}[0]{0.08}
\begin{figure*}[b]
    \centering
    \setlength\extrarowheight{2pt}
    \begin{tabular}{ccccccccc}
    Image 1 &&&&&&&& Image 2 \\
    \includegraphics[height=\hb\linewidth,width=\wb\linewidth,keepaspectratio]{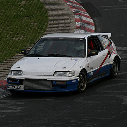} &
    \includegraphics[height=\hb\linewidth,width=\wb\linewidth,keepaspectratio]{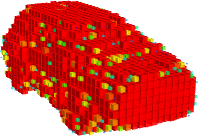} &
    \includegraphics[height=\hb\linewidth,width=\wb\linewidth,keepaspectratio]{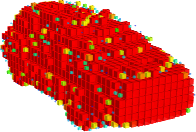} &
    \includegraphics[height=\hb\linewidth,width=\wb\linewidth,keepaspectratio]{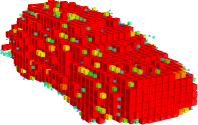} &
    \includegraphics[height=\hb\linewidth,width=\wb\linewidth,keepaspectratio]{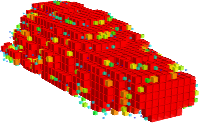} &
    \includegraphics[height=\hb\linewidth,width=\wb\linewidth,keepaspectratio]{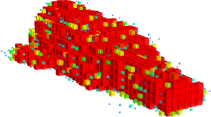} &
    \includegraphics[height=\hb\linewidth,width=\wb\linewidth,keepaspectratio]{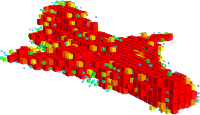} &
    \includegraphics[height=\hb\linewidth,width=\wb\linewidth,keepaspectratio]{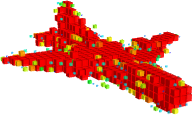} &
    \includegraphics[height=\hb\linewidth,width=\wb\linewidth,keepaspectratio]{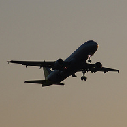} \\
    \includegraphics[height=\hb\linewidth,width=\wb\linewidth,keepaspectratio]{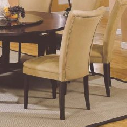} &
    \includegraphics[height=\hb\linewidth,width=\wb\linewidth,keepaspectratio]{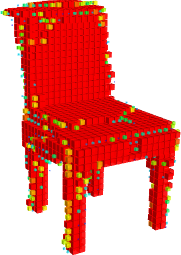} &
    \includegraphics[height=\hb\linewidth,width=\wb\linewidth,keepaspectratio]{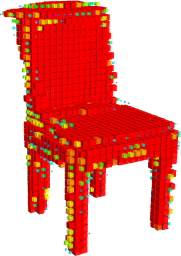} &
    \includegraphics[height=\hb\linewidth,width=\wb\linewidth,keepaspectratio]{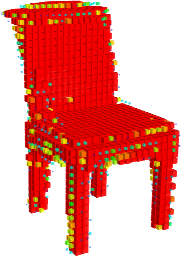} &
    \includegraphics[height=\hb\linewidth,width=\wb\linewidth,keepaspectratio]{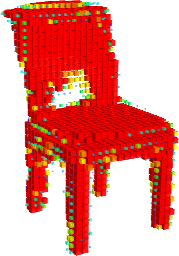} &
    \includegraphics[height=\hb\linewidth,width=\wb\linewidth,keepaspectratio]{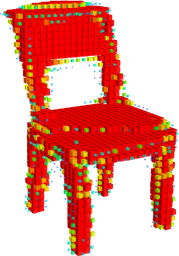} &
    \includegraphics[height=\hb\linewidth,width=\wb\linewidth,keepaspectratio]{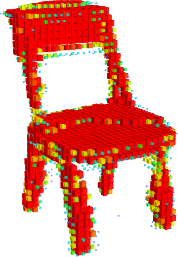} &
    \includegraphics[height=\hb\linewidth,width=\wb\linewidth,keepaspectratio]{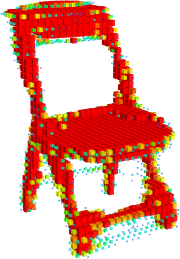} &
    \includegraphics[height=\hb\linewidth,width=\wb\linewidth,keepaspectratio]{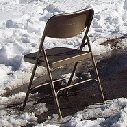} \\
    \includegraphics[height=0.08\linewidth,width=0.08\linewidth,keepaspectratio]{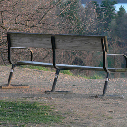} &
    \includegraphics[height=0.08\linewidth,width=0.08\linewidth,keepaspectratio]{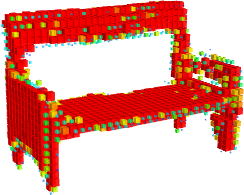} &
    \includegraphics[height=0.08\linewidth,width=0.08\linewidth,keepaspectratio]{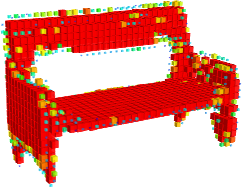} &
    \includegraphics[height=0.08\linewidth,width=0.08\linewidth,keepaspectratio]{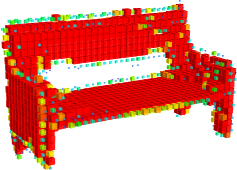} &
    \includegraphics[height=0.08\linewidth,width=0.08\linewidth,keepaspectratio]{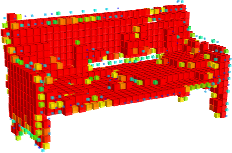} &
    \includegraphics[height=0.08\linewidth,width=0.08\linewidth,keepaspectratio]{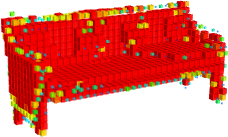} &
    \includegraphics[height=0.08\linewidth,width=0.08\linewidth,keepaspectratio]{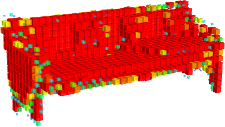} &
    \includegraphics[height=0.08\linewidth,width=0.08\linewidth,keepaspectratio]{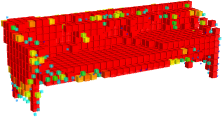} &
    \includegraphics[height=0.08\linewidth,width=0.08\linewidth,keepaspectratio]{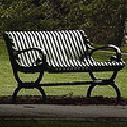} \\
    \includegraphics[height=0.08\linewidth,width=0.08\linewidth,keepaspectratio]{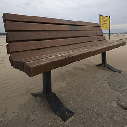} &
    \includegraphics[height=0.08\linewidth,width=0.08\linewidth,keepaspectratio]{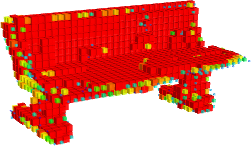} &
    \includegraphics[height=0.08\linewidth,width=0.08\linewidth,keepaspectratio]{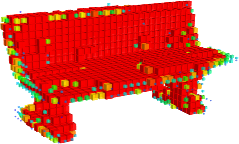} &
    \includegraphics[height=0.08\linewidth,width=0.08\linewidth,keepaspectratio]{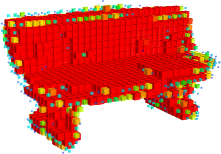} &
    \includegraphics[height=0.08\linewidth,width=0.08\linewidth,keepaspectratio]{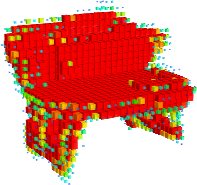} &
    \includegraphics[height=0.08\linewidth,width=0.08\linewidth,keepaspectratio]{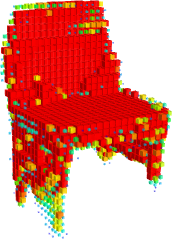} &
    \includegraphics[height=0.08\linewidth,width=0.08\linewidth,keepaspectratio]{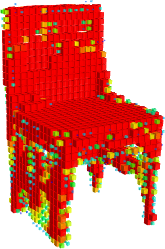} &
    \includegraphics[height=0.08\linewidth,width=0.08\linewidth,keepaspectratio]{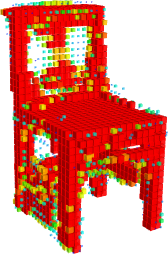} &
    \includegraphics[height=0.08\linewidth,width=0.08\linewidth,keepaspectratio]{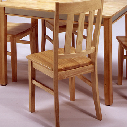} \\
    \includegraphics[height=0.08\linewidth,width=0.08\linewidth,keepaspectratio]{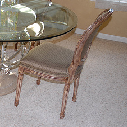} &
    \includegraphics[height=0.08\linewidth,width=0.08\linewidth,keepaspectratio]{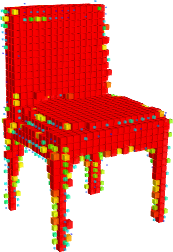} &
    \includegraphics[height=0.08\linewidth,width=0.08\linewidth,keepaspectratio]{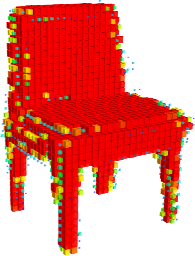} &
    \includegraphics[height=0.08\linewidth,width=0.08\linewidth,keepaspectratio]{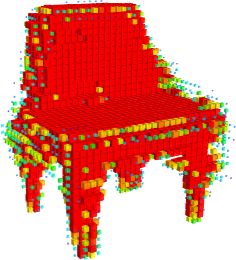} &
    \includegraphics[height=0.08\linewidth,width=0.08\linewidth,keepaspectratio]{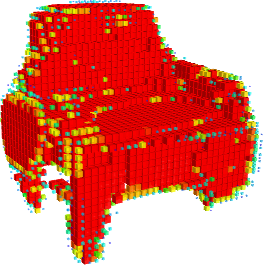} &
    \includegraphics[height=0.08\linewidth,width=0.08\linewidth,keepaspectratio]{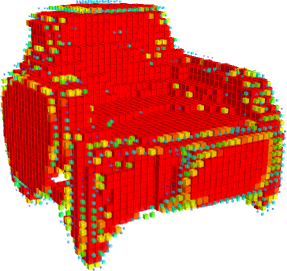} &
    \includegraphics[height=0.08\linewidth,width=0.08\linewidth,keepaspectratio]{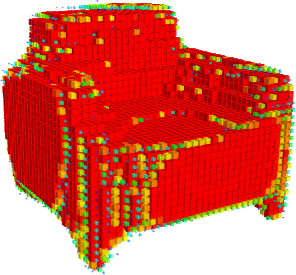} &
    \includegraphics[height=0.08\linewidth,width=0.08\linewidth,keepaspectratio]{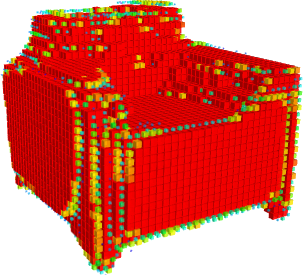} &
    \includegraphics[height=0.08\linewidth,width=0.08\linewidth,keepaspectratio]{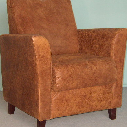} \\
    \includegraphics[height=0.08\linewidth,width=0.08\linewidth,keepaspectratio]{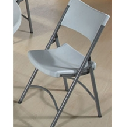} &
    \includegraphics[height=0.08\linewidth,width=0.08\linewidth,keepaspectratio]{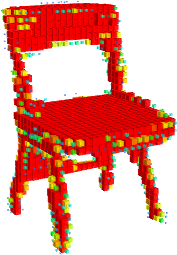} &
    \includegraphics[height=0.08\linewidth,width=0.08\linewidth,keepaspectratio]{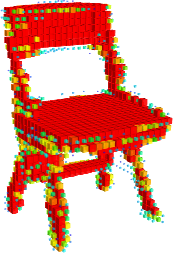} &
    \includegraphics[height=0.08\linewidth,width=0.08\linewidth,keepaspectratio]{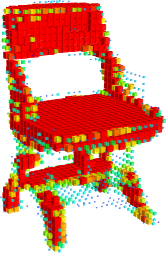} &
    \includegraphics[height=0.08\linewidth,width=0.08\linewidth,keepaspectratio]{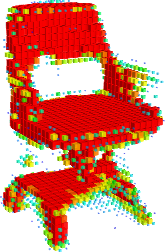} &
    \includegraphics[height=0.08\linewidth,width=0.08\linewidth,keepaspectratio]{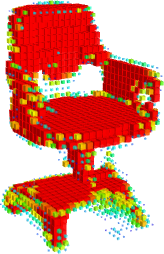} &
    \includegraphics[height=0.08\linewidth,width=0.08\linewidth,keepaspectratio]{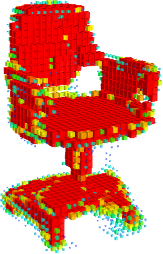} &
    \includegraphics[height=0.08\linewidth,width=0.08\linewidth,keepaspectratio]{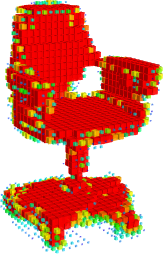} &
    \includegraphics[height=0.08\linewidth,width=0.08\linewidth,keepaspectratio]{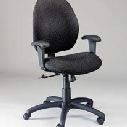} \\
    \includegraphics[height=0.08\linewidth,width=0.08\linewidth,keepaspectratio]{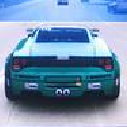} &
    \includegraphics[height=0.08\linewidth,width=0.08\linewidth,keepaspectratio]{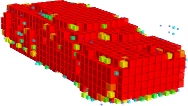} &
    \includegraphics[height=0.08\linewidth,width=0.08\linewidth,keepaspectratio]{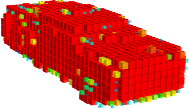} &
    \includegraphics[height=0.08\linewidth,width=0.08\linewidth,keepaspectratio]{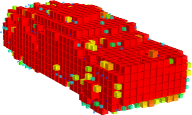} &
    \includegraphics[height=0.08\linewidth,width=0.08\linewidth,keepaspectratio]{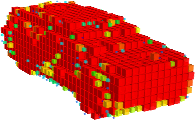} &
    \includegraphics[height=0.08\linewidth,width=0.08\linewidth,keepaspectratio]{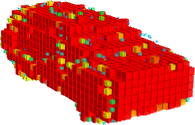} &
    \includegraphics[height=0.08\linewidth,width=0.08\linewidth,keepaspectratio]{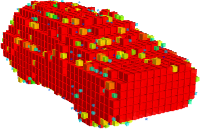} &
    \includegraphics[height=0.08\linewidth,width=0.08\linewidth,keepaspectratio]{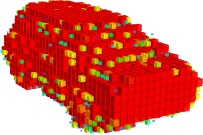} &
    \includegraphics[height=0.08\linewidth,width=0.08\linewidth,keepaspectratio]{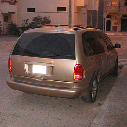} \\
    \includegraphics[height=0.08\linewidth,width=0.08\linewidth,keepaspectratio]{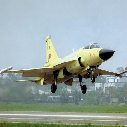} &
    \includegraphics[height=0.08\linewidth,width=0.08\linewidth,keepaspectratio]{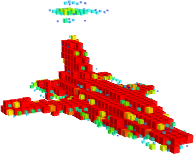} &
    \includegraphics[height=0.08\linewidth,width=0.08\linewidth,keepaspectratio]{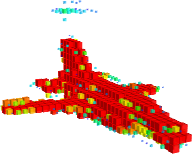} &
    \includegraphics[height=0.08\linewidth,width=0.08\linewidth,keepaspectratio]{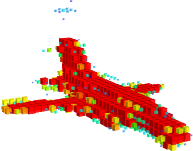} &
    \includegraphics[height=0.08\linewidth,width=0.08\linewidth,keepaspectratio]{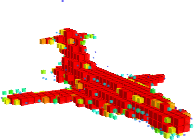} &
    \includegraphics[height=0.08\linewidth,width=0.08\linewidth,keepaspectratio]{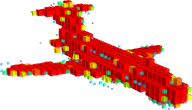} &
    \includegraphics[height=0.08\linewidth,width=0.08\linewidth,keepaspectratio]{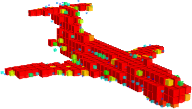} &
    \includegraphics[height=0.08\linewidth,width=0.08\linewidth,keepaspectratio]{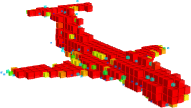} &
    \includegraphics[height=0.08\linewidth,width=0.08\linewidth,keepaspectratio]{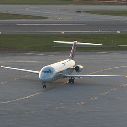} \\
    \includegraphics[height=0.08\linewidth,width=0.08\linewidth,keepaspectratio]{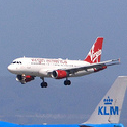} &
    \includegraphics[height=0.08\linewidth,width=0.08\linewidth,keepaspectratio]{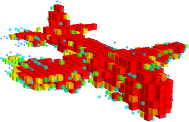} &
    \includegraphics[height=0.08\linewidth,width=0.08\linewidth,keepaspectratio]{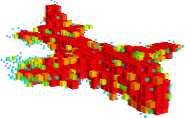} &
    \includegraphics[height=0.08\linewidth,width=0.08\linewidth,keepaspectratio]{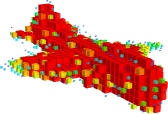} &
    \includegraphics[height=0.08\linewidth,width=0.08\linewidth,keepaspectratio]{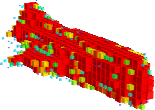} &
    \includegraphics[height=0.08\linewidth,width=0.08\linewidth,keepaspectratio]{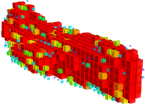} &
    \includegraphics[height=0.08\linewidth,width=0.08\linewidth,keepaspectratio]{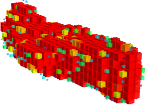} &
    \includegraphics[height=0.08\linewidth,width=0.08\linewidth,keepaspectratio]{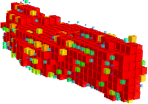} &
    \includegraphics[height=0.08\linewidth,width=0.08\linewidth,keepaspectratio]{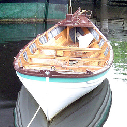} \\
    \end{tabular}
    \caption{Linear interpolation of latent variable $z$. We observed the smooth transition of objects inter-and intra-class. Interestingly, semantic properties of the object, such as the length of the airplane wings and the size of the hole in the back of the chair smoothly transitioned. This result hints that our network generalized such semantic properties in the latent variable $z$.}
    \label{fig:interpolation-supp}
\end{figure*}

